\documentclass[10pt,twocolumn,letterpaper]{article}

\PassOptionsToPackage{table}{xcolor}

\usepackage{cvpr}              

\definecolor{dark_green}{rgb}{0, 0.5, 0}

\definecolor{cvprblue}{rgb}{0.21,0.49,0.74}
\usepackage[pagebackref,breaklinks,colorlinks,allcolors=cvprblue]{hyperref}

\usepackage{amsmath,amsfonts,bm}

\def\eqref#1{equation~\ref{#1}}

\def\1{\bm{1}}

\DeclareMathAlphabet{\mathsfit}{\encodingdefault}{\sfdefault}{m}{sl}
\SetMathAlphabet{\mathsfit}{bold}{\encodingdefault}{\sfdefault}{bx}{n}

\def\1n{\mathbf{1}_n}
\def\0{\mathbf{0}}
\def\1{\mathbf{1}}

\definecolor{pink}{rgb}{0.9,0.5,0.5}
\definecolor{purple}{rgb}{0.5, 0.4, 0.8}   
\definecolor{gray}{rgb}{0.3, 0.3, 0.3}
\definecolor{mygreen}{rgb}{0.2, 0.6, 0.2}

\definecolor{greena}{rgb}{0.4, 0.5, 0.1}

\definecolor{bluea}{rgb}{0, 0.4, 0.6}

\definecolor{reda}{rgb}{0.6, 0.2, 0.1}

\newcommand{\cm}[1]{}

\newcommand{\myheading}[1]{\vspace{1ex}\noindent \textbf{#1}}

\newif\ifshowsolution
\showsolutiontrue

\ifshowsolution  

\else

\fi

\makeatletter
\DeclareRobustCommand\onedot{\futurelet\@let@token\@onedot}
\def\@onedot{\ifx\@let@token.\else.\null\fi\xspace}

\def\Approach{PDTrack\xspace}

\definecolor{graybg}{gray}{0.75} 

\definecolor{custom_red}{RGB}{231,111,81}
\definecolor{custom_green}{RGB}{42,157,143}
\definecolor{custom_dark}{RGB}{38,70,83}
\definecolor{custom_yellow}{RGB}{233,196,106}
\definecolor{custom_orange}{RGB}{244,162,97}

\usepackage{url}
\usepackage{array}    
\usepackage{multirow} 
\usepackage{tabularx}
\usepackage{hhline}
\usepackage{booktabs}
\usepackage{subcaption}

\usepackage{graphicx}
\usepackage{arydshln}
\usepackage{multirow}
\usepackage{comment}
\usepackage[table]{xcolor}
\usepackage{xspace}

\usepackage{threeparttable} 

\usepackage{amsmath,amsfonts,bm}

\usepackage{makecell}

\definecolor{custom_red}{RGB}{231,111,81}
\definecolor{custom_green}{RGB}{42,157,143}
\definecolor{custom_dark}{RGB}{38,70,83}
\definecolor{custom_yellow}{RGB}{233,196,106}
\definecolor{custom_orange}{RGB}{244,162,97}

\def\paperID{3272} 
\def\paperID{xxxx} 
\def\confName{CVPR}
\def\confYear{2026}

\def\Approach{DAGE\xspace}

\title{\Approach: Dual-Stream Architecture for \\Efficient and Fine-Grained Geometry Estimation}

\author{Tuan Duc Ngo\textsuperscript{1,}\footnotemark[2]
\hspace*{1em} 
Jiahui Huang\textsuperscript{2}
\hspace*{1em} 
Seoung Wug Oh\textsuperscript{2}
\hspace*{1em} 
Kevin Blackburn-Matzen\textsuperscript{2}
\\ 
Evangelos Kalogerakis\textsuperscript{1,3}
\hspace*{1em}
Chuang Gan\textsuperscript{1}
\hspace*{1em} 
Joon-Young Lee\textsuperscript{2}
\\
\textsuperscript{1} UMass Amherst \, \textsuperscript{2} Adobe Research \,
\textsuperscript{3} TU Crete \\
\href{https://ngoductuanlhp.github.io/dage-site/}{github.com/dage-site}
}

\begin{document}

\twocolumn[{%
\renewcommand\twocolumn[1][]{#1}%
\maketitle
\begin{center}
\vspace{-8mm}
\includegraphics[width=0.93\linewidth]{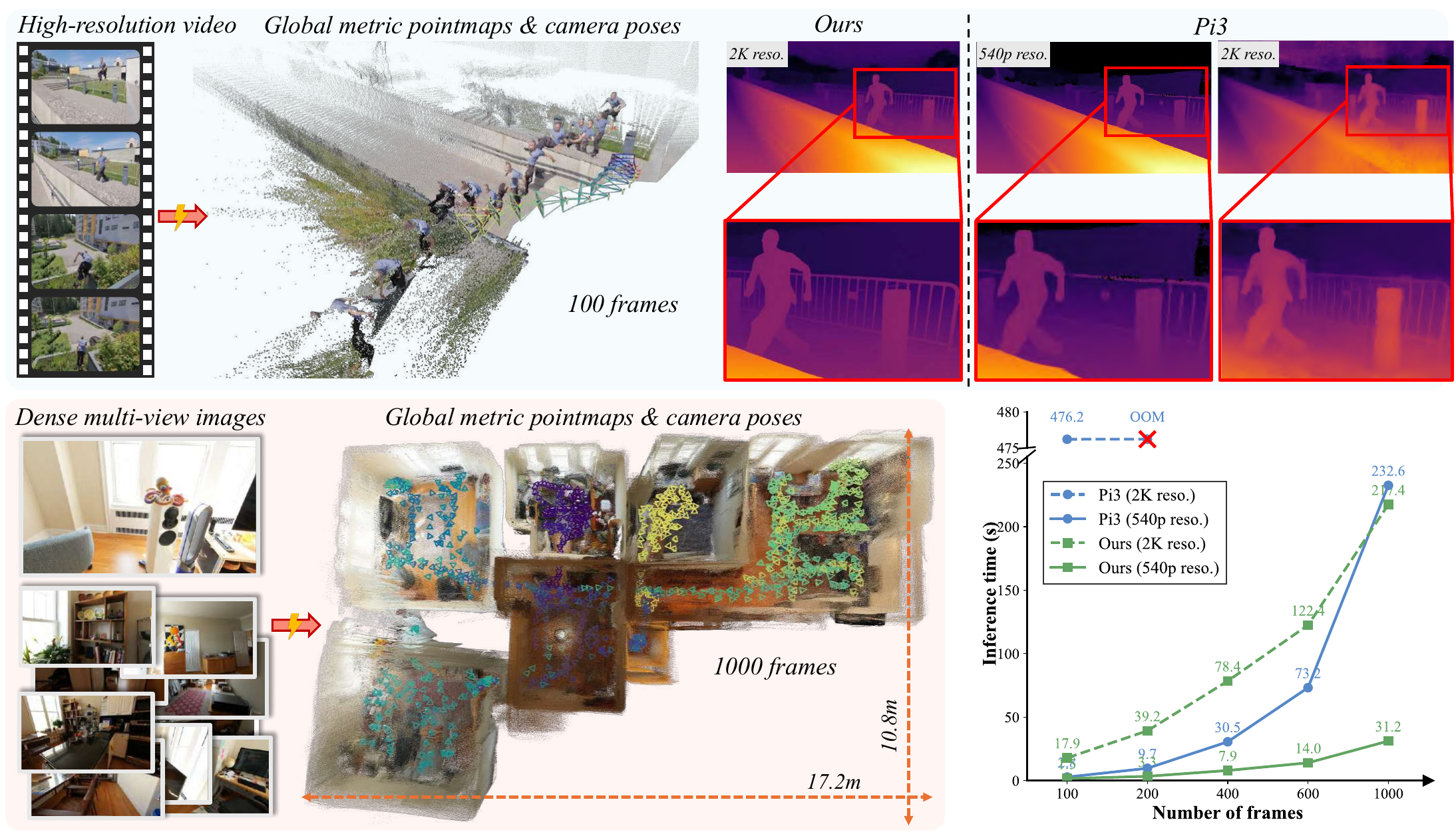}
\vspace{-3mm}
\captionof{figure}{\textbf{\Approach} produces \emph{high-resolution, fine-grained, metric-scale} and \emph{cross-view consistent} 3D geometry together with accurate camera poses from visual inputs. It runs substantially faster than prior models~\cite{pi3,vggt} and scales to long sequences (up to 1000 frames).}

\label{fig:teaser}
\end{center}
}]

\renewcommand{\thefootnote}{} 
\footnotetext{$^{\dagger}$ work done during internship at Adobe Research.}
\renewcommand{\thefootnote}{\arabic{footnote}} 

\begin{abstract}
Estimating accurate, view-consistent geometry and camera poses from uncalibrated multi-view/video inputs remains challenging—especially at high spatial resolutions and over long sequences. We present \textbf{\Approach}, a dual-stream transformer whose main novelty is to disentangle global coherence from fine detail. A \textbf{low-resolution} stream operates on aggressively downsampled frames with alternating frame/global attention to build a view-consistent representation and estimate cameras efficiently, while a \textbf{high-resolution} stream processes the original images per-frame to preserve sharp boundaries and small structures. A lightweight \textbf{adapter} fuses these streams via cross-attention, injecting global context without disturbing the pretrained single-frame pathway. This design scales resolution and clip length \emph{independently}, supports inputs up to 2K, and maintains practical inference cost. \Approach delivers sharp depth/pointmaps, strong cross-view consistency, and accurate poses, establishing new state-of-the-art results for video geometry estimation and multi-view reconstruction.
\end{abstract} 
\vspace{-6mm}
\section{Introduction}
\label{sec:intro}



Estimating 3D geometry and camera poses from multi-view images is a fundamental problem in computer vision. We target the demanding regime of \emph{uncalibrated, high-resolution} inputs with potentially \emph{thousands of frames}. This task is particularly challenging, as the model must simultaneously (i) enforce global consistency across views, (ii) preserve fine-grained details at high resolution, and (iii) remain tractable in runtime and memory for long sequences.


On one hand, feed-forward visual geometry networks~\cite{mvdust3r, fast3r, flare, cut3r, vggt, pi3,mapanything} have achieved remarkable progress in globally consistent multi-view geometry estimation, setting new state-of-the-art results on various benchmarks~\cite{e3d_bench,benchmarking_stereo}, including video depth estimation, 3D reconstruction, and camera pose prediction. However, their typically heavy network architectures limit training and inference to modest image resolutions (e.g., long side $ \leq $ 518px) and a small number of input views, which leads to blurred thin structures and poorly defined object boundaries. Several works have adopted post-training acceleration strategies~\cite{fastervggt,fastvggt,quantvggt} to reduce computational cost and support more views during inference, yet they do not address the loss of high-frequency details or the tendency toward oversmoothed surfaces near edges and small objects.

On the other hand, single-view geometry estimators \cite{moge,moge2,depthanything,depthanythingv2,depthpro} operate flexibly at high resolution and produce sharp, detail-rich depth/pointmaps from single images, yet they lack temporal and multi-view consistency by design. Attempts to adapt these models to handle videos \cite{rollingdepth,Khan2023TemporallyCO, geometrycrafter,tracktention,stereodiff, Cho2025SeuratFM} introduce heavy pipelines, and typically do not recover accurate camera poses. As a result, they fail to assemble a globally consistent 3D scene geometry directly from the feed-forward predictions. 

Based on this observation, we present \Approach, a \textbf{D}ual-stream \textbf{A}rchitecture for efficient and fine-grained \textbf{G}eometry \textbf{E}stimation that meets the above criteria. It comprises two parallel streams and a lightweight fusion adapter. The \textbf{Low-Resolution (LR) Stream} focuses on extracting globally consistent features and predicting camera poses. It is composed of a ViT backbone followed by a global transformer with alternating frame-global attention~\cite{vggt,pi3}, which processes the entire sequence at a lower spatial resolution. Although the global transformer is computationally intensive, operating at low resolution keeps it tractable while preserving global context.
The \textbf{High-Resolution (HR) Stream} is designed to capture high-frequency details and fine-grained features. It employs a ViT~\cite{vit} that processes each image independently at its native resolution.  Finally, our proposed \textbf{Lightweight Adapter} synchronizes and fuses LR and HR tokens before the dense heads, yielding geometry that is both globally consistent and richly detailed.

This decoupled design grants two critical advantages. \emph{First, it achieves global consistency and tractability.} By restricting the computationally-heavy global attention to the LR stream, we alleviate the quadratic scaling bottleneck of global transformers~\cite{vggt,pi3}. This significantly reduces runtime, by $2\times$ and $28\times$ at 540p and 2K resolutions, respectively, enabling our model to process thousands of frames. \emph{Second, it preserves high-fidelity detail.} The HR stream operates per-frame, allowing it to scale to any resolution (e.g., up to 2K) and leverage priors from state-of-the-art single-image models for sharp detail and strong real-world generalization. In contrast to standard pipelines~\cite{vggt,pi3,flare,fast3r} that couple image resolution with sequence length, \Approach\ decouples the two, enabling independent control over spatial detail and multi-view coherence, with a tractable runtime (see Fig.~\ref{fig:teaser}).

We validate our method and design choices through extensive experiments. \Approach achieves state-of-the-art performance on video geometry and depth-sharpness benchmarks, and is competitive on 3D reconstruction and camera pose estimation—while offering higher throughput and a lower GPU memory footprint. In summary, our technical contributions are twofold:
\begin{itemize}
    \item A \emph{dual-stream transformer} that couples a per-frame, high-resolution detail path with a multi-view, low-resolution global-attention path.
    \item A lightweight \emph{Adapter} that fuses the two streams to produce sharp yet globally consistent geometry.
\end{itemize}





\section{Related Work}
\label{sec:related_work}

\begin{figure*}[t]
\begin{center}
\includegraphics[width=0.95\linewidth]{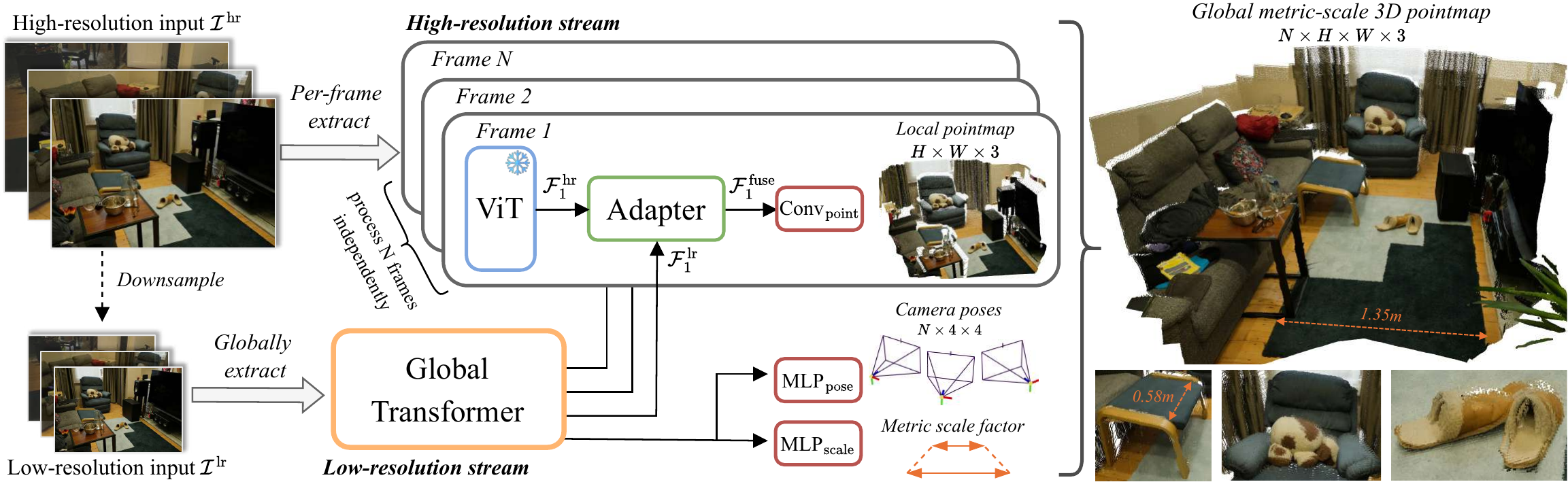}

\vspace{-2mm}
\caption{\textbf{Overview of \Approach.} Given a set of \emph{unposed} RGB images, the model predicts per-frame pointmaps and camera poses, plus a scene-wise metric scale. The architecture has two parallel streams: (i) a low-resolution stream (lower part) that processes downsampled inputs to aggregate global context and regress poses/scene scale; and (ii) a high-resolution stream (upper part) that processes frames independently at native resolution to preserve fine detail. A lightweight Adapter fuses LR and HR tokens before the dense geometry head.}

\vspace{-8mm}
\label{fig:arch}
\end{center}
\end{figure*}

\myheading{Single-view Geometry Estimation} aims to recover 3D scene geometry from a single image. Early approaches relied on handcrafted features and probabilistic models \cite{hoiem2007recovering,saxena2005learning,6787109,saxena2008make3d}. With deep networks, numerous architectures were proposed \cite{wang2018learning,fu2018deep,bhat2021adabins,bts,eigen2014depth}, yet their generalization remained limited. The introduction of relative-depth \cite{ranftl2020towards} enabled training on large, mixed datasets, leading to strong zero-shot performance \cite{zoedepth,metric3d,metric3dv2,unidepth,Guizilini2023TowardsZS,depthanything,depthanythingv2,depthpro}. However, many such methods require camera intrinsics and metric scale to recover absolute 3D geometry. Recent work addresses this by jointly estimating depth and intrinsics \cite{leres,unidepth,depthpro}. A complementary line of research regresses dense \emph{3D pointmaps} directly \cite{moge,moge2}, from which depths and intrinsics can be recovered. Despite impressive single-image results, these methods typically exhibit temporal jitter and inconsistent scale when applied to videos.

\myheading{Fine-Grained Geometry Estimation} targets predicting sharper depth/pointmaps with high-frequency detail. Patchwise fusion methods \cite{boosting_mono,patchfusion} enhance local detail by combining per-patch estimates, but often introduce stitching artifacts at patch boundaries. Another line of work leverages powerful generative priors \cite{stable_diffusion,stable_diffusion_xl} to produce highly detailed depth \cite{marigold,lotus,geowizard,finetune_diffusion,sharpdepth}. Depth Anything~V2 \cite{depthanythingv2} improves detail via large-scale, high-quality synthetic data, while DepthPro \cite{depthpro} employs a multi-patch ViT design \cite{vit} to better capture fine structures. MoGe2 \cite{moge2} combines synthetic and refined real-world annotations with a coarse-to-fine loss \cite{moge}, achieving strong metric accuracy and sharp predictions. \textit{Concurrently}, \cite{pixel_perfect_depth} integrates foundation-model geometry priors~\cite{depthanythingv2,vggt} with a cascaded DiT~\cite{dit}, yielding \emph{pixel-perfect depth}.
Nonetheless, these methods are predominantly per-image and fail to ensure temporal consistency in video setting.


\myheading{Video-based Geometry Estimation} mitigates temporal jitter and scale inconsistency by stabilizing \emph{per-frame} predictions with test-time procedures or using video architectures. Several works regularize single-image depth across time using geometric consistency or online refinement \cite{lai2018learning,luo2020consistent,kopf2021robust,droidslam}, or optimize scale/shift to co-align frames \cite{rollingdepth}. \cite{video_depth_anything,flashdepth,tracktention} add temporal heads or video transformers on top of pretrained single-view models, and diffusion-based pipelines~\cite{svd,ho2022imagen,cogvideox} leverage strong video priors for temporally consistent depth. Despite impressive performance, diffusion-based methods are compute-intensive and typically do not recover camera poses.

\myheading{Visual Geometry Estimation} regresses both camera poses and the 3D scene structure from uncalibrated images or videos. Classical SfM/MVS pipelines \cite{snavely2006photo,agarwal2011building,wu2013towards,schonberger2016structure}, are robust but require multi-stage optimization. ~\cite{leapvo,delta,uni4d} inject motion priors (e.g. optical flow, point tracking) and then perform bundle adjustment, which reduces manual engineering but requires per-video optimization. Dust3R \cite{dust3r} introduced a learning-based alternative that predicts pointmaps for image pairs in a shared coordinate frame and stitches multi-view input via a global alignment step. Subsequent work improves metric-scale recovery \cite{mast3r}, extends to dynamic scenes \cite{monst3r,align3r,easi3r,st4rtrack}, and scales multi-view processing \cite{mvdust3r,dens3r,fast3r,flare,cut3r,vggt,pi3,mapanything,point3r,streamvggt}. Among these, VGGT \cite{vggt} and Pi3 \cite{pi3} demonstrate state-of-the-art performance with alternating global-frame attention transformers. However, the quadratic cost of global attention imposes tight token budgets, limiting input resolution and the number of frames; thus, predicted depth often appears blurred and fine structures are smoothed.

In contrast, our dual-stream approach performs feed-forward global aggregation on LR inputs for efficiency while preserving HR detail via a per-frame stream, with a lightweight adapter to fuse these two.

\section{Method}
\label{sec:method}

\subsection{Problem Definition}
\label{sec:problem}

Given an \emph{uncalibrated} set of $N$ RGB images $\mathcal{I}=\{I_i\}_{i=1}^{N}$ of a scene, where each $I_i \in \mathbb{R}^{H\times W\times 3}$, our model aims to reconstruct the 3D scene geometry by predicting three components:
(1) per-frame pointmaps $\mathcal{P}=\{P_i\}_{i=1}^{N}$, where $P_i \in \mathbb{R}^{H\times W\times 3}$ represents the 3D coordinates of each pixel in the local camera coordinate system;
(2) camera poses $\mathcal{G}=\{G_i\}_{i=1}^{N}$, where $G_i \in \mathrm{SE}(3)$ encodes each camera’s rotation and translation; and
(3) a single global metric scale factor $s \in \mathbb{R}^+$.


State-of-the-art feed-forward approaches~\cite{vggt,pi3} are constrained by the high computational cost of global attention. This typically limits their inputs to modest resolutions (e.g., 518px on the long side) and short sequences (e.g., $N < 200$). 
We address this limitation with a dual-stream architecture designed to produce high-quality, fine-grained 3D geometry and accurate camera poses while supporting flexible spatial resolutions and long sequences.


Fig.~\ref{fig:arch} illustrates the overall architecture of our model. It consists of a \emph{low-resolution (LR) stream} (Sec.\ref{sec:lr_stream}) and a \emph{high-resolution (HR) stream} (Sec.\ref{sec:hr_stream}), which operate in parallel and are synchronized through a lightweight adapter (Sec.\ref{sec:adapter}), followed by dense prediction heads (Sec.\ref{sec:prediction_heads}). The LR stream extracts globally consistent features and estimates camera poses, while the HR stream predicts per-frame pointmaps at the native input resolution. The global features produced by the LR stream are injected into the HR stream via the adapter to enhance geometric consistency across views. Training details are described in Sec.\ref{sec:training}.


\begin{figure}[t]
\begin{center}
\includegraphics[width=1\linewidth]{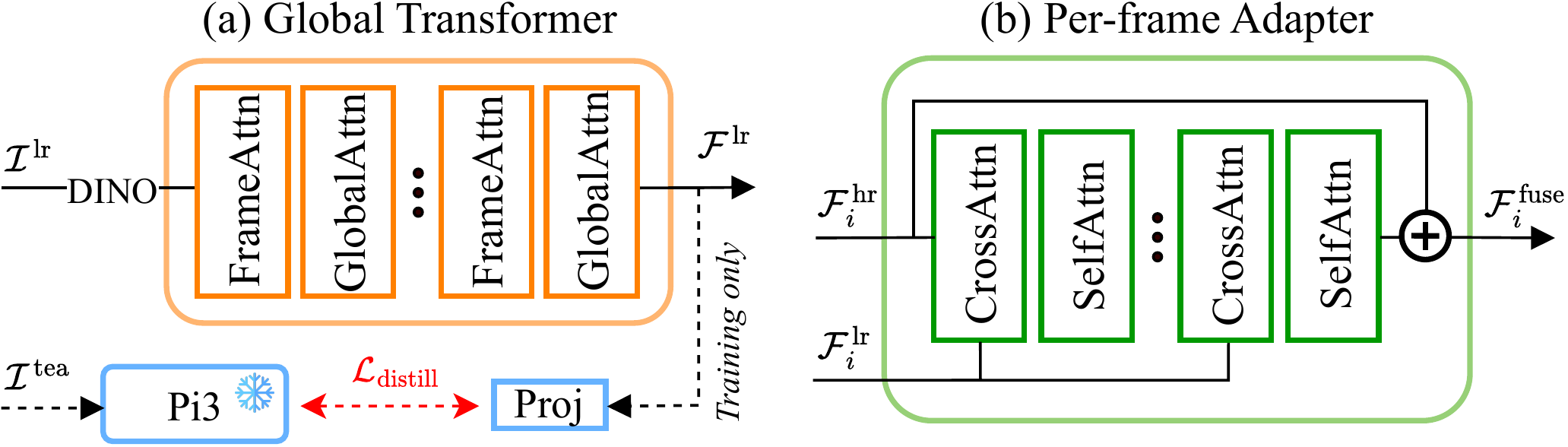}
\vspace{-6mm}
\caption{The \textbf{Global transformer} (left) operates on low-resolution inputs with alternating global and frame-wise attention; during training, feature distillation compensates for aggressive downsampling. The \textbf{Adapter} (right) stacks cross and self-attention blocks to fuse multi-view–consistent LR tokens into the HR stream.}
\vspace{-8mm}
\label{fig:video_trans_adapter}
\end{center}
\end{figure}

\subsection{Low-Resolution Stream}
\label{sec:lr_stream}
The low-resolution stream is responsible for enforcing global consistency and estimating camera poses. It processes the entire sequence $\{I_i\}$ at a fixed low resolution (long side $\leq 252$px), $\mathcal{I}^{\mathrm{lr}}$. These images are passed through a global transformer to output \emph{LR feature tokens} $\mathcal{F}^{\mathrm{lr}} \in \mathbb{R}^{N\times h_{lr}\times w_{lr}\times C}$. This global transformer (Fig.~\ref{fig:video_trans_adapter}a) consists of a DINOv2~\cite{dinov2} tokenizer, and alternating blocks of frame-wise and global self-attention (\texttt{[FrameAttn $\rightarrow$ GlobalAttn]})~\cite{vggt,pi3}, which is effective for capturing scene-level structure. We do not use dedicated camera tokens, preserving permutation equivariance~\cite{pi3}.


While low-resolution processing ensures tractability, training the LR stream from scratch often leads to degraded camera pose accuracy. To address this, we leverage the rich global representations of a pre-trained teacher model, Pi3~\cite{pi3}, through knowledge distillation. Specifically, the teacher processes a higher-resolution input $\mathcal{I}^{\mathrm{tea}}$ (capped at 518px to match~\cite{pi3,vggt}) and produces features $\mathcal{F}^{\mathrm{tea}} \in \mathbb{R}^{N\times h_{\mathrm{tea}}\times w_{\mathrm{tea}}\times C}$. These features are then used to supervise the LR stream via a feature distillation loss:

\vspace{-2mm}
\begin{equation}
\label{eq:distill}
\mathcal{L}_{\text{dis}} = 1-\text{sim}(p_{\phi}(\mathcal{F}^{\mathrm{lr}}), \mathcal{F}^{\mathrm{tea}})
\vspace{-1mm}
\end{equation}
\noindent where $\text{sim}(\cdot, \cdot)$ is the cosine similarity function, and $p_{\phi}$ is a projection network mapping student features to the teacher's representation space and spatial dimension.

\subsection{High-Resolution Stream}
\label{sec:hr_stream}
The high-resolution stream processes each frame of the input sequence $\{I_i\}$ independently at its original resolution. To preserve fine-grained detail and strong zero-shot generalization capabilities, we adopt MoGe2~\cite{moge2} as the HR backbone. This model uses a 24-layer ViT encoder~\cite{vit} to extract the \emph{HR feature tokens} $\mathcal{F}^{\mathrm{hr}} \in \mathbb{R}^{N\times h_{hr}\times w_{hr}\times C_{hr}}$.

\subsection{Adapter}
\label{sec:adapter}
The adapter (Fig.~\ref{fig:video_trans_adapter}b) is designed to inject global context from the LR stream into the per-frame HR stream. A naive solution—such as upsampling LR features via interpolation and concatenating them with HR features—often introduces artifacts and fails to capture meaningful cross-view relations. Alternative approaches, including pixel-shuffle upsampling~\cite{pixel_perfect_depth, hourglass_dit} or CNN-based upsampling~\cite{lift}, alleviate such artifacts but rely on a fixed scale factor, which is too restrictive for inputs that may vary up to 2K resolution.

To overcome these limitations, we adopt a more flexible \emph{cross-attention} mechanism that accommodates arbitrary token counts from both streams. This fusion is followed by HR self-attention to restore per-frame spatial coherence. Concretely, for each $i$-th frame, the fused feature $\mathcal{F}^{\mathrm{fuse}}_{i}$ is computed as:
\vspace{-2mm}
\begin{align}
\vspace{-2mm}
\mathcal{F}^{\mathrm{fuse}}_i &= \mathrm{CrossAttn}\!\big(Q=\mathcal{F}^{\mathrm{hr}}_i;\, K,V=\mathcal{F}^{\mathrm{lr}}_i\big)\\
\mathcal{F}^{\mathrm{fuse}}_i  &= \mathcal{F}^{\mathrm{hr}}_i + \mathrm{MLP}\!\big(\mathrm{SelfAttn}(\mathcal{F}^{\mathrm{fuse}}_i)\big)
\end{align}
Positional encodings are applied before attention to align HR patch coordinates with their LR counterparts, stabilizing the cross-scale fusion. We stack five such \texttt{[CrossAttn $\rightarrow$ SelfAttn]} blocks.

We employ Rotary Positional Encodings (RoPE) for all attention layers, but with different strategies for self-attention and cross-attention to handle varying resolutions. \textbf{Self-Attention:} Standard RoPE does not extrapolate well to spatial dimensions larger than those seen during training, which can cause distortion on high-resolution inputs~\cite{cut3r,vggt,pi3}. Thus, we adopt the \emph{interpolated RoPE}~\cite{rope_interp} technique. We define a fixed maximum patch length, $l^{\mathrm{max}}$. At both training and inference, we rescale the angular frequencies of the positional encoding to this fixed context size, which keeps the positional spectrum stable even at very high resolutions.
\textbf{Cross-Attention:} A challenge is the large spatial mismatch between the LR and HR streams (e.g., 252px vs. 2K). To align them, we ``snap'' each HR token to its nearest grid cell in the LR feature map. The HR token then uses the positional encoding from that corresponding LR cell. This simple strategy effectively matches patches across scales and avoids extrapolation, as the LR stream's spatial dimensions are always fixed and within the trained bounds. Concretely, let $R(\mathbf{f}, \mathbf{m})$ be the RoPE function applied to token $\mathbf{f}$ at 2D position $\mathbf{m}$. The modified RoPE functions are:
\vspace{-4mm}
\begin{align}
\vspace{-2mm}
R_{\text{self}}\!\big(\mathbf{f}^{\mathrm{hr}}, \mathbf{m}^{\mathrm{hr}}\big)
  &= R\!\left(\mathbf{f}^{\mathrm{hr}},\, \mathbf{m}^{\mathrm{hr}} \cdot l^{\mathrm{max}}/l^{\mathrm{hr}}\right),\\
R_{\text{cross}}\!\big(\mathbf{f}^{\mathrm{hr}}, \mathbf{m}^{\mathrm{hr}}\big)
  &= R\!\left(\mathbf{f}^{\mathrm{hr}},\, \mathrm{sampling}\!\left(\mathbf{m}^{\mathrm{hr}},\, \mathrm{grid}^{\mathrm{lr}}\right)\right)
\end{align}
\noindent
where $l^{\mathrm{hr}}$ is the side length of the HR grid.




\noindent
\textbf{Adapter Design Discussion.} We investigated various strategies for fusing the LR and HR tokens, focusing on \textit{where} and \textit{how} to inject the global information. 
For \textit{where} to insert, one approach is to inject intermediate LR features into each ViT layer of the HR stream. This mitigates scale drift but fails to enforce cross-view global consistency (see Sec.~\ref{sec:ablation}).
For \textit{how} to fuse, we considered alternatives like concatenation and addition with learnable interpolation. We find that the best trade-off is a lightweight adapter after the HR ViT encoder, comprising cross-attention to inject global context and self-attention to re-calibrate intra-frame coherence (see Sec.~\ref{sec:ablation}). This strategy preserves the HR stream's original feature space at the start of training, allowing the model to gradually learn to incorporate the multi-view consistent constraints to refine the final geometry.

\subsection{Prediction Heads}
\label{sec:prediction_heads}
\myheading{Dense Geometry.} We employ a feature pyramid of convolutional layers~\cite{moge2} to gradually upsample the per-patch features $\mathcal{F}^{\mathrm{fuse}}$ into dense feature maps at the original image resolution to regress the pointmaps $\mathcal{P}$. This convolutional-style head yields smoother predictions, avoiding the grid-like artifacts observed in~\cite{pi3} (see Fig.~\ref{fig:quali_recons}).

\vspace{-1mm}
\myheading{Camera Pose.} We regress the per-frame camera parameters using the LR features $\mathcal{F}^{\mathrm{lr}}$. This is done for efficiency, as camera poses do not require fine-grained features. Following~\cite{reloc3r,pi3}, we use average pooling and an MLP to regress the translation and rotation in a 9D representation~\cite{Levinson2020AnAO}.

\vspace{-1mm}
\myheading{Metric Scale.} We add a \emph{metric scale} token in the video transformer, followed by an MLP to predict a single metric scale factor for each scene.

\begin{table*}[t]
\centering
\caption{\textbf{Video pointmap evaluation}. Results are aligned with the ground truth by optimizing a shared scale and shift factor across the entire video. “MV/HR/PO” indicate multi-view support, high-resolution input support, and whether the method predicts camera poses. We mark {\setlength{\fboxsep}{2pt}\colorbox{custom_green!70}{\strut best}} and {\setlength{\fboxsep}{2pt}\colorbox{custom_green!20}{\strut second-best}}.}
\vspace{-3mm}
\small
\setlength{\tabcolsep}{1pt}
\renewcommand{\arraystretch}{1.15}
\resizebox{\textwidth}{!}{%
\begin{threeparttable}
\begin{tabular}{l|ccc|cc|cc|cc|cc|cc|cc|cc|cc|c}
\toprule
\multirow{3}{*}{Method} & \multirow{3}{*}{MV} & \multirow{3}{*}{HR} & \multirow{3}{*}{PO} &
\multicolumn{2}{c}{\textbf{GMU}~\cite{gmu_kitchen}} &
\multicolumn{2}{c}{\textbf{Monkaa}~\cite{monkaa}} &
\multicolumn{2}{c}{\textbf{Sintel}~\cite{sintel}} &
\multicolumn{2}{c}{\textbf{ScanNet}~\cite{scannet}} &
\multicolumn{2}{c}{\textbf{KITTI}~\cite{kitti}} &
\multicolumn{2}{c}{\textbf{UrbanSyn}~\cite{urbansyn}} &
\multicolumn{2}{c}{\textbf{Unreal4K}~\cite{unreal4k}} &
\multicolumn{2}{c}{\textbf{Diode}~\cite{diode}} &
\multirow{3}{*}{\textbf{Rank} $\downarrow$}
\\
& & & & \multicolumn{2}{c|}{($960\times512$)} & \multicolumn{2}{c|}{($960\times512$)} & \multicolumn{2}{c|}{($896\times448$)} & \multicolumn{2}{c|}{($640\times512$)} & \multicolumn{2}{c|}{($768\times384$)} & \multicolumn{2}{c|}{($2048\times1024$)} & \multicolumn{2}{c|}{($1920\times1080$)} & \multicolumn{2}{c|}{($1024\times768$)} \\
& & & &
$\mathrm{Rel}^{p}\!\downarrow$ & $\delta^{p}\!\uparrow$ &
$\mathrm{Rel}^{p}\!\downarrow$ & $\delta^{p}\!\uparrow$ &
$\mathrm{Rel}^{p}\!\downarrow$ & $\delta^{p}\!\uparrow$ &
$\mathrm{Rel}^{p}\!\downarrow$ & $\delta^{p}\!\uparrow$ &
$\mathrm{Rel}^{p}\!\downarrow$ & $\delta^{p}\!\uparrow$ &
$\mathrm{Rel}^{p}\!\downarrow$ & $\delta^{p}\!\uparrow$ &
$\mathrm{Rel}^{p}\!\downarrow$ & $\delta^{p}\!\uparrow$ &
$\mathrm{Rel}^{p}\!\downarrow$ & $\delta^{p}\!\uparrow$ & \\
\midrule
DepthPro~\cite{depthpro}            &   & \checkmark &   &
9.5 & 93.9 &
25.1 & 58.4 &
40.8 & 44.7 &
9.3 & 94.9 &
10.0 & 94.9 &
48.9 & 40.1 &
74.7 & 12.0 &
32.4 & 59.2 &
7.9 \\
MoGe~\cite{moge}                    &   & \checkmark &   &
20.3 & 71.2 &
22.9 & 61.3 &
29.4 & 59.8 &
13.4 & 88.0 &
8.0 & 95.8 &
14.9 & 87.0 &
38.3 & 51.5 &
31.8 & 52.9  &
7.4 \\
MoGe2~\cite{moge2}                  &   & \checkmark &   &
19.6 & 72.4 &
25.0 & 57.0 &
29.8 & 58.4 &
12.4 & 89.4 &
9.0 & 97.2 &
13.4 & 90.0 &
32.9 & 59.1 &
31.0 & 54.2 &
6.8 \\
MoGe2~\cite{moge2}$^{\dagger}$      & $\triangle$  & \checkmark &   &
7.1 & \cellcolor{custom_green!20}94.6 &
21.4 & 67.6 &
28.2 & 62.8 &
7.8 & 97.5 &
10.5 & 98.4 &
\cellcolor{custom_green!70}7.2 & \cellcolor{custom_green!70}97.1 
&
12.6 & 86.7 &
15.8 & 84.1 &
4.1 \\
CUT3R~\cite{cut3r}                  & \checkmark &   & \checkmark &
8.0 & 93.7 &
31.8 & 47.5 &
35.8 & 47.5 &
5.9 & 97.9 &
14.5 & 87.5 &
21.6 & 68.6 &
16.8 & 79.6 &
17.9 & 78.3 &
6.8 \\
VGGT~\cite{vggt}                    & \checkmark &   & \checkmark &
5.4 & 93.8 &
13.6 & 84.4 &
23.7 & \cellcolor{custom_green!20}73.1 &
2.9 & 99.0 &
7.5 & 97.4 &
14.5 & 87.3 &
\cellcolor{custom_green!70}8.6 & \cellcolor{custom_green!70}96.1 &
13.3 & 85.9 &
3.4 \\
Pi3~\cite{pi3}                      & \checkmark &   & \checkmark &
\cellcolor{custom_green!20}5.2 & 94.2 &
\cellcolor{custom_green!20}11.6 & \cellcolor{custom_green!20}90.0 &
\cellcolor{custom_green!20}22.0 & 72.9 &
\cellcolor{custom_green!20}2.2 & \cellcolor{custom_green!20}99.4 &
6.3 & 97.3 &
16.8 & 77.5 &
19.5 & 75.3 &
\cellcolor{custom_green!70}9.2 & \cellcolor{custom_green!70}95.2 &
\cellcolor{custom_green!20}3.3 \\
GeoCrafter~\cite{geometrycrafter}   & \checkmark & $\triangle$  &   &
8.3 & \cellcolor{custom_green!70}94.8 &
15.7 & 83.4 &
25.0 & 69.3 &
8.3 & 96.9 &
\cellcolor{custom_green!70}5.6 & \cellcolor{custom_green!20}98.8 &
12.5 & 91.9 &
21.6 & 74.5 &
12.5 & 93.0 &
3.9 \\
\cdashline{1-21}
\textbf{\Approach (ours)}                        & \checkmark & \checkmark & \checkmark &
\cellcolor{custom_green!70}4.9 & 94.2 &
\cellcolor{custom_green!70}10.1 & \cellcolor{custom_green!70}91.0 &
\cellcolor{custom_green!70}21.5 & \cellcolor{custom_green!70}75.6 &
\cellcolor{custom_green!70}2.1 & \cellcolor{custom_green!70}99.5 &
\cellcolor{custom_green!20}5.9 & \cellcolor{custom_green!70}99.0 &
\cellcolor{custom_green!20}8.8 & \cellcolor{custom_green!20}96.0 &
\cellcolor{custom_green!20}11.9 & \cellcolor{custom_green!20}89.1 &
\cellcolor{custom_green!20}9.7 & \cellcolor{custom_green!20}94.4 &
\cellcolor{custom_green!70}1.6 \\   
\bottomrule
\end{tabular}
\begin{tablenotes}\footnotesize
\item $\triangle$: \textit{partial support.} \quad $^{\dagger}$: obtained by multiplying per-frame pointmap by predicted metric scale factor.
\end{tablenotes}
\end{threeparttable}
} 
\vspace{-4mm}
\label{tab:quanti_video_pointmap}
\end{table*}

\subsection{Training Details}
\label{sec:training}

\subsubsection{Training loss}

We train the model with a combination of pointmap, camera, scale, normal, gradient, and distillation losses.

\myheading{Pointmap loss.}
We predict a per-pixel 3D point $\hat{\mathbf{p}}_{i,j}$ up to a scene-wide scale. Let
$\mathrm{norm}(\cdot)$ denote a scene normalization (distance-to-origin) applied to both prediction and ground truth. We compute a single alignment scale $s^{*}$ using the ROE solver~\cite{moge} and supervise with an $\ell_{1}$ loss:
\vspace{-2mm}
\begin{equation}
\mathcal{L}_{\text{pm}}
= \frac{1}{NHW}\sum_{i=1}^{N}\sum_{j=1}^{H\!\times\! W}
\left\lVert
\,\frac{s^{*}\hat{\mathbf{p}}_{i,j}}{\mathrm{norm}(\hat{\mathcal{P}})}
-
\frac{\mathbf{p}_{i,j}}{\mathrm{norm}(\mathcal{P})}
\right\rVert_{1}.
\label{eq:l_points}
\vspace{-1mm}
\end{equation}
Unlike uncertainty-weighted objectives used in \cite{dust3r,vggt}, we do \emph{not} attenuate errors with confidences, as we found this can suppress hard structures and reduce sharpness.


\myheading{Camera loss.}
Following \cite{pi3}, we supervise \emph{relative} camera poses to avoid fixing a global coordinate frame. Let
$\hat{\mathbf{g}}_{uv}$ and $\mathbf{g}_{uv}$ denote predicted and ground-truth pairwise poses between frames $u$-th and $v$-th, the camera loss is defined as:
\vspace{-2mm}
\begin{equation}
\vspace{-2mm}
\mathcal{L}_{\text{camera}}
= \frac{1}{N(N-1)}
\sum_{\substack{u,v=1\\ u\neq v}}^{N}
\mathcal{L}_{\text{cam}}\!\big(\hat{\mathbf{g}}_{uv},\,\mathbf{g}_{uv}\big),
\label{eq:l_camera}
\end{equation}
\noindent
where $\mathcal{L}_{\text{cam}}$ comprises $\mathcal{L}_{\text{rot}}$ that minimizes the geodesic distance of the rotation part, and $\mathcal{L}_{\text{trans}}$ is the $\ell_1$ loss of the translation part.

\myheading{Scale loss.}
For datasets with metric supervision, we additionally supervise the predicted metric scale $\hat{s}$:
\vspace{-2mm}
\begin{equation}
\vspace{-2mm}
\mathcal{L}_{\text{scale}}
= \left\lVert \log \hat{s} \;-\; \log\!\left( s^{*}\,\frac{\mathrm{norm}(\mathcal{P})}{\mathrm{norm}(\hat{\mathcal{P}})} \right) \right\rVert_{2}.
\end{equation}

\begin{figure*}
    \centering
    
    \includegraphics[width=0.95\linewidth]{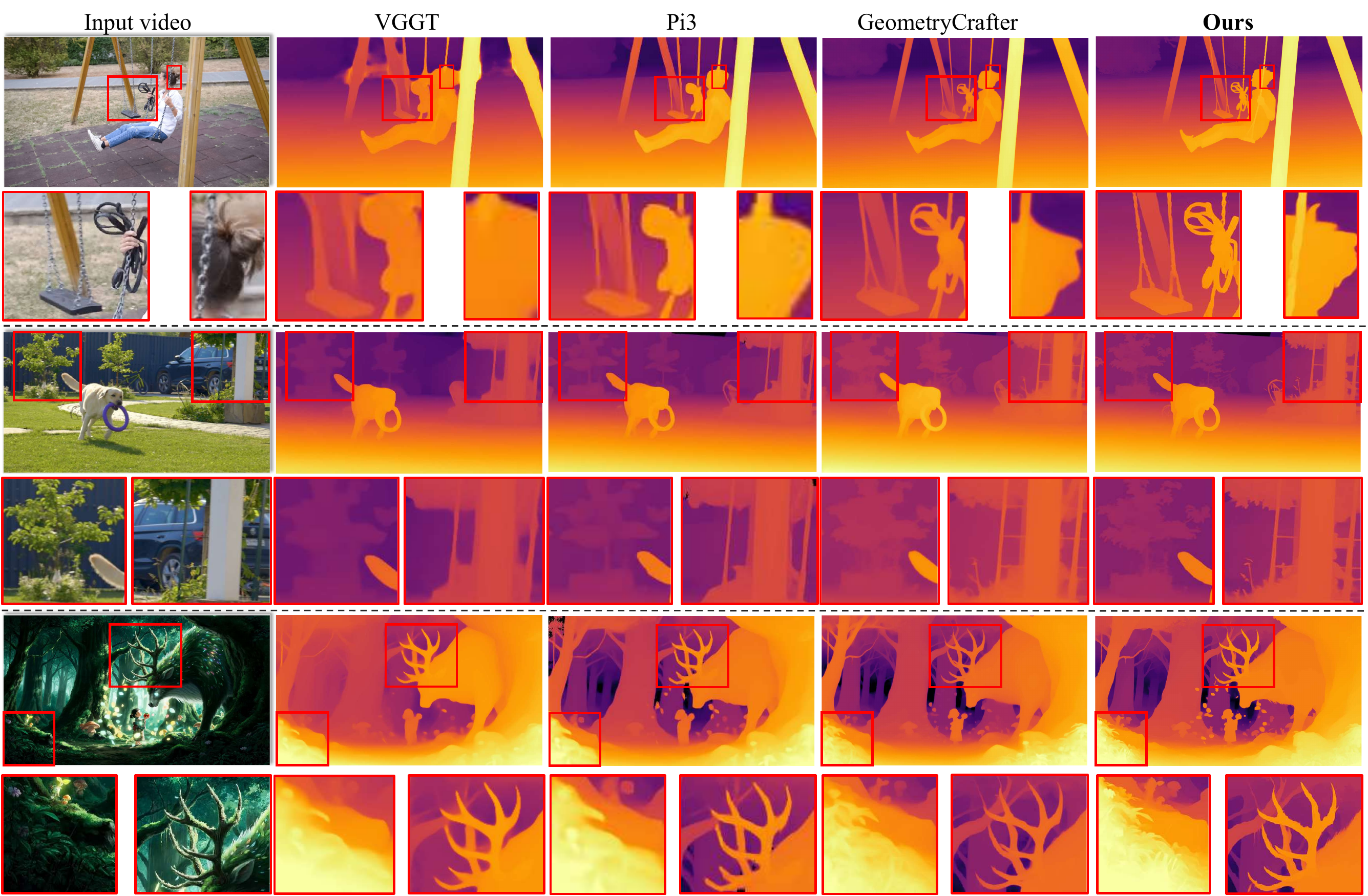}
    \vspace{-3mm}
    \caption{\textbf{Visual comparison of video depth on \emph{in-the-wild} scenes.} We convert the depth map to a disparity map for better visualization, and zoom-in (\textcolor[RGB]{255,0,0}{red bounding boxes}) to emphasize details. \Approach preserves sharp boundaries and fine-grained detail—especially for thin structures and small or distant objects, outperforming a diffusion-based baseline~\cite{geometrycrafter}.}
    \vspace{-5mm}
    \label{fig:quali_depth}
\end{figure*}

\myheading{Normal loss.}
To encourage locally smooth but faithful surfaces, we supervise normals computed \emph{on-the-fly} from the pointmap via cross products~\cite{moge}:
\vspace{-2mm}
\begin{equation}
\vspace{-2mm}
\mathcal{L}_{\text{normal}}
= \frac{1}{NHW}\sum_{i=1}^{N}\sum_{j=1}^{H\!\times\! W} \angle\!\big(\hat{\mathbf{n}}_{i,j},\,\mathbf{n}_{i,j}\big),
\label{eq:normal_loss}
\end{equation}
where $\angle(\cdot,\cdot)$ is the angular difference.

\myheading{Gradient loss.}
To improve local geometry, MoGe~\cite{moge} applies a \emph{multi-scale affine-invariant pointmap loss} by subsampling local regions at several scales and aligning each region to ground truth independently. While this improves single-image sharpness, we found that per-region independent alignments introduce patch-wise degrees of freedom that break cross-view consistency, leading to seams and drift—undesirable in our multi-view setting (see Tab.~\ref{tab:ablation_sharp}). Instead, we preserve a \emph{single global alignment} and encourage detail by supervising \emph{image gradients} of the canonical inverse depth~\cite{depthpro} at multiple scales:
\vspace{-2mm}
\begin{equation}
\vspace{-2mm}
\mathcal{L}_{\text{gradient}}
= \frac{1}{NHW}\sum_{i=1}^{N}\sum_{j=1}^{H\!\times\! W}
\left\lVert
\nabla\, \hat{d}_{i,j}
-
\nabla\, d_{i,j}
\right\rVert_{1},
\label{eq:l_edge}
\end{equation}
where $\hat{d}$ and $d$ denote ground-truth and predicted canonical inverse depth, respectively, and $\nabla$ is the Scharr and Laplace gradient filters. 

Due to the sparsity of real-world depth annotations, we only apply the normal and gradient loss on synthetic data.




\subsubsection{Implementation details}
The HR stream uses a frozen 24-layer ViT from MoGe2~\cite{moge2}. Since our training corpus is relatively small, we initialize the LR stream from Pi3~\cite{pi3} instead of training from scratch. The adapter comprises five blocks, each containing a cross-attention and self-attention layer. We train \Approach on 18 publicly available datasets spanning indoor$/$outdoor, static$/$dynamic scenes. The complete list of datasets and implementation details are provided in the supplementary.

\section{Experiments}
\label{sec:exp}

This section compares our method to state-of-the-art approaches across four tasks to show its effectiveness.
\subsection{Video Geometry Estimation}
We evaluate on eight datasets spanning diverse conditions—GMU Kitchens~\cite{gmu_kitchen}, ScanNet~\cite{scannet} (indoor RGB-D), KITTI~\cite{kitti} (outdoor driving with LiDAR), Sintel~\cite{sintel} and Monkaa~\cite{monkaa} (synthetic with precise depth and challenging dynamics), and the high-resolution UrbanSyn~\cite{urbansyn}, Unreal4K~\cite{unreal4k}, and Diode~\cite{diode}-and resolutions from $\sim$640p to 2K. Following~\cite{moge,geometrycrafter}, we report relative point error Rel$^{p}\downarrow$ and inlier ratio $\delta^{p}\uparrow$ at a 0.25 threshold, and evaluate \emph{affine-invariant} pointmaps by aligning predictions to ground truth with a single, shared scale and shift per video. We compare against single-image methods~\cite{depthpro,moge,moge2}, video diffusion-based model~\cite{geometrycrafter}, and set-based visual-geometry models~\cite{cut3r,vggt,pi3}. For methods that do not support high-resolution inference~\cite{cut3r,vggt,pi3, geometrycrafter}, inputs are downsampled to the model’s native resolution and outputs are upsampled to the original size to avoid degenerate predictions. In Tab.~\ref{tab:quanti_video_pointmap}, \Approach delivers consistently strong performance across datasets—achieving state-of-the-art average rank on Rel$^{p}$ and $\delta^{p}$—with pronounced gains on high-resolution scenarios.
We visualize the disparity map of our approach and other baselines in Fig.~\ref{fig:quali_depth}.
Detailed evaluations (including \textit{scale-invariant}, \textit{affine-invariant} and video depth estimation) are provided in the supplementary.

\begin{figure*}
    \centering
    \includegraphics[width=0.95\linewidth]{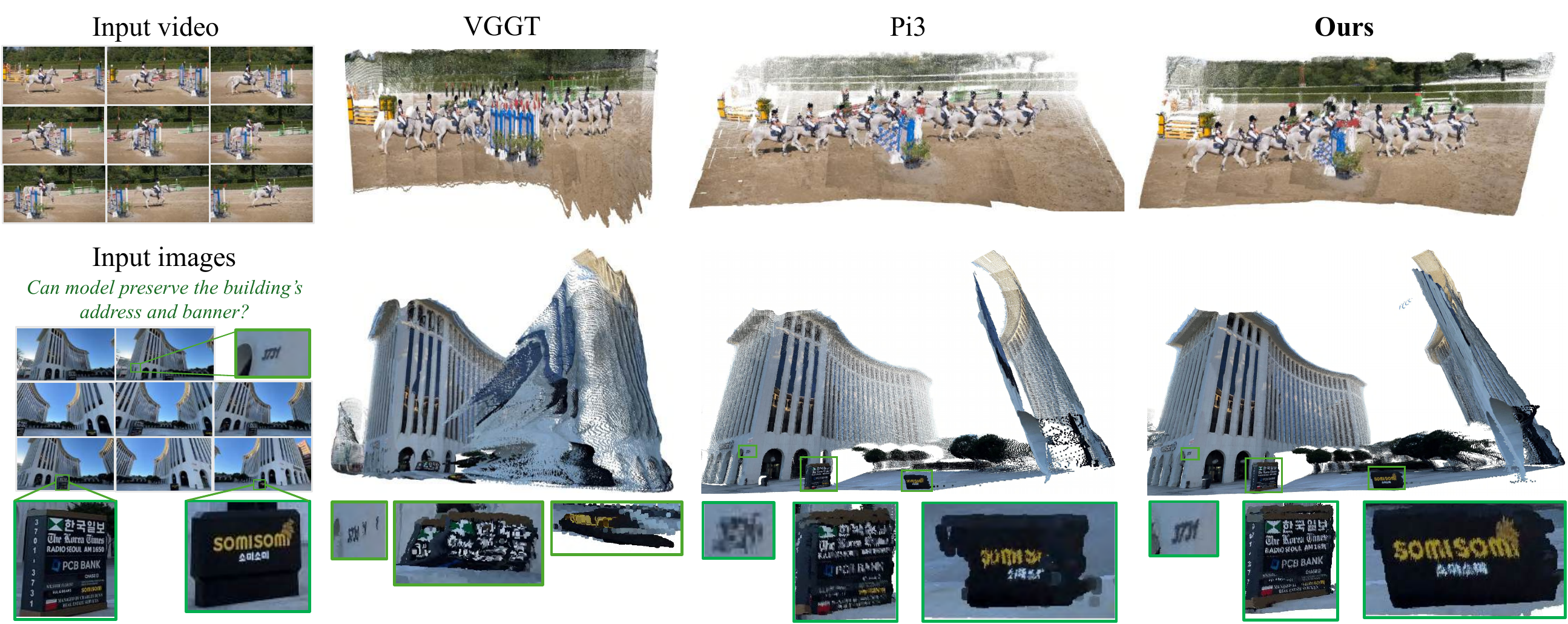}
    \vspace{-3mm}
    \caption{\textbf{Visual comparison of 3D reconstruction on \emph{in-the-wild} scenes.}
    Compared to VGGT~\cite{vggt} and Pi3~\cite{pi3}, \Approach achieves comparable multi-view consistency while preserving markedly finer detail (\textcolor[RGB]{78,167,46}{green boxes}). 
    }
    \vspace{-5mm}
    \label{fig:quali_recons}
\end{figure*}

\subsection{Sharp Depth Estimation}

We assess depth–boundary sharpness on four synthetic datasets with high-quality depth annotations —Monkaa~\cite{monkaa}, Sintel~\cite{sintel}, UrbanSyn~\cite{urbansyn}, and Unreal4K~\cite{unreal4k}. 
Following~\cite{depthpro}, we report the scale-invariant boundary F1$\uparrow$, which compares occlusion contours induced by neighboring-pixel depth ratios in prediction versus ground truth. Because F1 does not reflect temporal stability, we also measure the pseudo depth boundary error ($C_{\mathrm{PDBE}}\!\downarrow$)~\cite{sharpdepth}, defined as the Chamfer distance between prediction and ground-truth at canny-detected edges. For a fair comparison of sharpness detail, we evaluate methods at each dataset’s native resolution; models that run out of memory (e.g., \cite{geometrycrafter, vggt}) are downsampled to the largest feasible resolution. Results in Tab.~\ref{tab:quanti_sharpdepth} show that, among methods producing temporally consistent video depth~\cite{cut3r,pi3,vggt,geometrycrafter}, \Approach achieves the highest F1 and the lowest PDBE. While DepthPro~\cite{depthpro} attains a higher F1 on some datasets, \Approach yields lower $C_{\mathrm{PDBE}}$—indicating more temporally consistent boundaries in the video setting.

\subsection{Multi-view Reconstruction}
Following \cite{cut3r,pi3}, we evaluate reconstructed multi-view \emph{pointmaps} on 7-Scenes~\cite{7scenes} and NRGBD~\cite{nrgbd} under \emph{sparse} and \emph{dense} settings. 
Predictions are first aligned to ground truth via Umeyama \(\mathrm{Sim}(3)\), then refined with ICP. We report accuracy Acc.$\downarrow$, completeness Comp.$\downarrow$, and normal consistency NC$\uparrow$ in Tab.~\ref{tab:quanti_mvrecon}. Comparisons include recent feed-forward visual-geometry methods~\cite{fast3r,flare,cut3r,vggt,pi3,mapanything}. We also assess metric-scale reconstruction by aligning with rigid transformation \(\mathrm{SE}(3)\), comparing against metric-pointmap methods~\cite{cut3r,mapanything}. Across sparse and dense settings, \Approach matches state-of-the-art performance~\cite{vggt,pi3} while recovering metric-accurate geometry. Fig.~\ref{fig:quali_recons} shows that our model produces globally consistent pointmaps while preserving fine-grained details.

\subsection{Camera Pose Estimation}

We evaluate on the synthetic Sintel~\cite{sintel} and two real-world datasets, TUM-Dynamics~\cite{tum_dynamic} and ScanNet~\cite{scannet}. We report \textit{Absolute Trajectory Error} (ATE) and \textit{Relative Pose Error} for translation/rotation (RPE\(_T\)/RPE\(_R\)). Predicted camera trajectories are registered to ground truth with a \(\mathrm{Sim}(3)\) alignment. We summarize the performance in Tab.~\ref{tab:quanti_pose}. Notably, we run the LR stream at 252px (long side) to estimate poses efficiently. Competing methods~\cite{pi3,vggt,cut3r} typically require ~518px to achieve accurate predictions. Despite using lower resolution inputs, \Approach matches their performance at their high-res settings—and outperform them when evaluated at the same low-res setting.

\subsection{Runtime Comparison}
Tab.~\ref{tab:runtime} reports the FPS and GPU memory consumption averaged over ten 100-frame videos on a single A100 GPU. \Approach sustains 65.4 FPS at 540p, which is $2\times$ faster than Pi3, and remains tractable at 5.6 FPS on 2K, where global-attention baselines~\cite{vggt, pi3} often run out of memory (\textit{OOM}). It is consistently faster than multi-view methods~\cite{geometrycrafter,vggt,cut3r} and adds only marginal overhead over the single-view MoGe2~\cite{moge2}—thanks to the decoupled LR/HR design that confines heavy global attention to the LR path, keeping runtime largely insensitive to HR input size. 


\begin{table}[t]
\centering
\caption{\textbf{Sharpness depth evaluation}.}
\vspace{-3mm}
\resizebox{\linewidth}{!}{
\setlength{\tabcolsep}{1pt}
\begin{threeparttable}
    
\begin{tabular}{l | cc | cc | cc | cc}
\toprule
\multirow{2}{*}{Method} &
\multicolumn{2}{c|}{\textbf{Monkaa}~\cite{monkaa}} &
\multicolumn{2}{c|}{\textbf{Sintel}~\cite{sintel}} &
\multicolumn{2}{c|}{\textbf{UrbanSyn}~\cite{urbansyn}} &
\multicolumn{2}{c}{\textbf{Unreal4K}~\cite{unreal4k}} \\
& F1$\uparrow$ & $C_{\mathrm{PDBE}}\downarrow$ &
  F1$\uparrow$ & $C_{\mathrm{PDBE}}\downarrow$ &
  F1$\uparrow$ & $C_{\mathrm{PDBE}}\downarrow$ &
  F1$\uparrow$ & $C_{\mathrm{PDBE}}\downarrow$ \\
\midrule
\color{black!40} DepthPro~\cite{depthpro} & \color{black!40} 0.19 & \color{black!40} 21.3 & \color{black!40} 0.41 & \color{black!40} 17.0 & \color{black!40} 0.14 & \color{black!40} 12.5 & \color{black!40} 0.07 & \color{black!40} 116.4 \\
\color{black!40} MoGe2~\cite{moge2} & \color{black!40} 0.27 & \color{black!40} 11.6 & \color{black!40} 0.27 & \color{black!40} 10.1 & \color{black!40} 0.09 & \color{black!40} 19.1 & \color{black!40} 0.10 & \color{black!40} 35.2  \\
\cdashline{1-9}
GeoCrafter~\cite{geometrycrafter}                   & \cellcolor{custom_green!20}0.19 &  \cellcolor{custom_green!70}8.86 & \cellcolor{custom_green!20}0.28 & \cellcolor{custom_green!20}8.1 & \cellcolor{custom_green!20}0.08\textsuperscript{$\star$} & 33.2\textsuperscript{$\star$}    &  \cellcolor{custom_green!20}0.06\textsuperscript{$\star$}   & 41.4\textsuperscript{$\star$} \\
CUT3R~\cite{cut3r}  & 0.08 & 20.3 & 0.11 & 16.5 & 0.01 & 44.0 & 0.01 & 63.1 \\
VGGT~\cite{vggt}    & 0.14 & 11.1 & 0.20 & 9.6 & 0.02\textsuperscript{$\star$} & 42.0\textsuperscript{$\star$} & 0.03\textsuperscript{$\star$} & \cellcolor{custom_green!20}38.1\textsuperscript{$\star$} \\
Pi3~\cite{pi3}                & 0.14 & 12.7 & 0.20 &  \cellcolor{custom_green!20}8.1 &  0.01 & \cellcolor{custom_green!20}27.9 & 0.03 & 46.9 \\
\cdashline{1-9}
\textbf{\Approach (ours)}                 & \cellcolor{custom_green!70}0.29 &  \cellcolor{custom_green!20}10.1 & \cellcolor{custom_green!70}0.34 & \cellcolor{custom_green!70}7.8 & \cellcolor{custom_green!70}0.09 & \cellcolor{custom_green!70}17.8 & \cellcolor{custom_green!70}0.14 & \cellcolor{custom_green!70}33.1 \\
\bottomrule
\end{tabular}
\begin{tablenotes}\small
\item $^\star$: \textit{Input resolution downscaled to prevent out-of-memory (OOM)}.
\end{tablenotes}
\end{threeparttable}
}
\vspace{-3mm}
\label{tab:quanti_sharpdepth}
\end{table}

\begin{table}[t]
\centering 
\caption{\textbf{Multi-view reconstruction evaluation.} We report the \textit{median} values on 3 settings, including \emph{sparse}, \emph{dense}, and \emph{metric}.}
\vspace{-3mm}
{\scriptsize 
\setlength{\tabcolsep}{2pt}
\resizebox{\linewidth}{!}{
\begin{tabular}{lc | ccc | ccc }
\toprule
\multirow{2}{*}{Method} & 
\multirow{2}{*}{Setting} &
\multicolumn{3}{c|}{\textbf{7-Scenes}~\cite{7scenes}} &
\multicolumn{3}{c}{\textbf{NRGBD}~\cite{nrgbd}} \\
& & Acc.$\downarrow$ & Comp.$\downarrow$ & NC$\uparrow$ & Acc.$\downarrow$ & Comp.$\downarrow$ & NC$\uparrow$ \\
\midrule
Fast3R~\cite{fast3r}  & \multirow{6}{*}{\rotatebox[origin=c]{90}{\itshape sparse}} & 0.065 & 0.089 & 0.759 & 0.091 & 0.104 & 0.877 \\
CUT3R~\cite{cut3r}   &                                 & 0.049 & 0.051 & 0.805 & 0.041 & 0.031 & 0.968 \\
FLARE~\cite{flare}   &                                 & 0.057 & 0.107 & 0.780 & 0.024 & 0.025 & 0.988 \\
VGGT~\cite{vggt}   &                                 & \cellcolor{custom_green!70}0.025 & \cellcolor{custom_green!70}0.033 & \cellcolor{custom_green!20}0.845 & 0.029 & 0.038 & 0.981 \\
Pi3~\cite{pi3} &                               & 0.029 & 0.049 & 0.841 & \cellcolor{custom_green!70}0.015 & \cellcolor{custom_green!70}0.014 & \cellcolor{custom_green!70}0.992 \\
MapAny~\cite{mapanything} & & 0.053 & 0.064 & 0.83 & 0.064 & 0.058 & 0.946  \\
\cdashline{1-8}
\textbf{\Approach (ours)} &                              & \cellcolor{custom_green!20}0.027 & \cellcolor{custom_green!20}0.042 & \cellcolor{custom_green!70}0.846 & \cellcolor{custom_green!20}0.018 & \cellcolor{custom_green!20}0.016 & \cellcolor{custom_green!70}0.992 \\
\midrule
Fast3R~\cite{fast3r}  & \multirow{6}{*}{\rotatebox[origin=c]{90}{\itshape dense}}  & 0.017 & 0.018 & 0.725 & 0.030 & 0.016 & 0.934 \\
CUT3R~\cite{cut3r}   &                                  & 0.010 & \cellcolor{custom_green!70}0.008 & 0.764 & 0.037 & 0.017 & 0.953 \\
FLARE~\cite{flare}  &                                  & \cellcolor{custom_green!70}0.007 & 0.013 & 0.785 & 0.011 & 0.008 & 0.986 \\
VGGT~\cite{vggt} &                                  & \cellcolor{custom_green!20}0.008 & 0.012 & 0.760 & 0.010 & \cellcolor{custom_green!70}0.005 & \cellcolor{custom_green!70} 0.988 \\
Pi3~\cite{pi3} &                                & \cellcolor{custom_green!70}0.007 & 0.011 & \cellcolor{custom_green!20}0.792 & \cellcolor{custom_green!70}0.008 & \cellcolor{custom_green!70}0.005 & \cellcolor{custom_green!20}0.987 \\
MapAny~\cite{mapanything} & & \cellcolor{custom_green!20}0.008 & \cellcolor{custom_green!70} 0.008 & 0.780 & 0.018 & 0.010 & 0.970 \\
\cdashline{1-8}
\textbf{\Approach (ours)} & & \cellcolor{custom_green!70}0.007 & \cellcolor{custom_green!20} 0.009 & \cellcolor{custom_green!70}0.793 & \cellcolor{custom_green!20}0.009 & \cellcolor{custom_green!20}0.006 & 0.985 \\ 
\midrule
CUT3R~\cite{cut3r} & \multirow{3}{*}{\rotatebox[origin=c]{90}{\itshape metric}} & \cellcolor{custom_green!20}0.189 & 0.186 & 0.582 & 0.307 & 0.253 & 0.606 \\
MapAny~\cite{mapanything} & & 0.339 & \cellcolor{custom_green!20}0.109 & \cellcolor{custom_green!20}0.639 & \cellcolor{custom_green!20}0.156 & \cellcolor{custom_green!20}0.108 & \cellcolor{custom_green!20}0.910 \\
\textbf{\Approach (ours)} & & \cellcolor{custom_green!70}0.034 & \cellcolor{custom_green!70}0.041 & \cellcolor{custom_green!70}0.847 & \cellcolor{custom_green!70}0.085 & \cellcolor{custom_green!70}0.101 & \cellcolor{custom_green!70}0.923  \\
\bottomrule
\end{tabular}
}
}
\vspace{-3mm}
\label{tab:quanti_mvrecon}
\end{table}

\begin{table}[t]
\centering
\caption{\textbf{Camera pose evaluation}}
\vspace{-3mm}
{\scriptsize 
\setlength{\tabcolsep}{1pt}
\resizebox{\linewidth}{!}{
\begin{tabular}{@{}l ccc ccc ccc@{}}
\toprule
\multirow{2}{*}{Method} &
\multicolumn{3}{c}{\textbf{Sintel}~\cite{sintel}} &
\multicolumn{3}{c}{\textbf{TUM-dynamics}~\cite{tum_dynamic}} &
\multicolumn{3}{c}{\textbf{ScanNet}~\cite{scannet}} \\
& ATE$\downarrow$ & RPE$_T$$\downarrow$ & RPE$_R$$\downarrow$ &
  ATE$\downarrow$ & RPE$_T$$\downarrow$ & RPE$_R$$\downarrow$ &
  ATE$\downarrow$ & RPE$_T$$\downarrow$ & RPE$_R$$\downarrow$ \\
\midrule
Fast3R~\cite{fast3r}      & 0.371 & 0.298 & 13.75 & 0.090 & 0.101 & 1.425 & 0.155 & 0.123 & 3.491 \\
CUT3R~\cite{cut3r}       & 0.217 & 0.070 & 0.636 & 0.047 & 0.015 & 0.451 & 0.094 & 0.022 & 0.629 \\
FLARE~\cite{flare} & 0.207 & 0.090 & 3.015 & 0.026 & 0.013 & 0.475 & 0.064 & 0.023 & 0.971 \\
VGGT~\cite{vggt}       & 0.167 & 0.062 & 0.491 & \cellcolor{custom_green!70}0.012 & \cellcolor{custom_green!20}0.010 & \cellcolor{custom_green!70}0.311 & 0.035 & 0.015 & \cellcolor{custom_green!20}0.382 \\
Pi3~\cite{pi3} & \cellcolor{custom_green!70} \cellcolor{custom_green!70}0.074 & \cellcolor{custom_green!70}0.040 & \cellcolor{custom_green!70}0.282 & \cellcolor{custom_green!20}0.014 & \cellcolor{custom_green!70}0.009 & \cellcolor{custom_green!20}0.312 & \cellcolor{custom_green!70}0.031 & \cellcolor{custom_green!70}0.013 & \cellcolor{custom_green!70}0.347 \\
\midrule
VGGT (252px) & 0.228 & 0.095 & 1.03 & 0.053 & 0.028 & 0.652 & 0.109 & 0.039 & 1.357 \\
Pi3 (252px) & 0.153 & 0.088 &	0.684 &	0.025 &	0.019 &	0.370 & 0.045 & 0.017 & 0.438 \\
\textbf{\Approach (ours)} & \cellcolor{custom_green!20}0.132 & \cellcolor{custom_green!20}0.051 & \cellcolor{custom_green!20}0.406 & \cellcolor{custom_green!20}0.014 & \cellcolor{custom_green!20}0.010 & 0.323 & \cellcolor{custom_green!70}0.031 & \cellcolor{custom_green!20}0.014 & 0.389 \\
\bottomrule
\end{tabular}
}
}
\vspace{-2mm}
\label{tab:quanti_pose}
\end{table}

\begin{table}[t]
\centering
\caption{\textbf{Throughput comparison.} FPS$\uparrow$ / GPU memory$\downarrow$ (GB) measured on 100-frame clips per resolution.}
\vspace{-3mm}
\scriptsize
\setlength{\tabcolsep}{1pt}
\begin{threeparttable}
\resizebox{\linewidth}{!}{
\begin{tabular}{l|*{6}{cc}}
\toprule
\multirow{2}{*}{Resolution} &
\multicolumn{2}{c}{\color{black!40}MoGe2$^\dag$} &
\multicolumn{2}{c}{GeoCrafter} &
\multicolumn{2}{c}{CUT3R} &
\multicolumn{2}{c}{VGGT} &
\multicolumn{2}{c}{Pi3} &
\multicolumn{2}{c}{\textbf{\Approach}} \\
&  {\color{black!40}FPS} & {\color{black!40}Mem.} & FPS & Mem. & FPS & Mem. & FPS & Mem. & FPS & Mem. & FPS & Mem. \\
\midrule
540$\times$360   & \color{black!40}79.4 & \color{black!40}8.1 & 3.1  & 17.3 & 27.2 & \cellcolor{custom_green!20}16.5 & 30.1 & 17.3  & \cellcolor{custom_green!20}36.3 & 17.2 & \cellcolor{custom_green!70}65.4 & \cellcolor{custom_green!70}10.1 \\
960$\times$512   &  \color{black!40}30.0 & \color{black!40}15.3 & 1.7  & 24.1 & \cellcolor{custom_green!20}20.3 & \cellcolor{custom_green!20}19.0 & 2.1  & 26.9 & 3.1  & 23.1 & \cellcolor{custom_green!70}28.9 & \cellcolor{custom_green!70}18.3 \\
2048$\times$1024 &  \color{black!40}6.1  & \color{black!40}22.1 & \multicolumn{2}{c}{\textit{OOM}} & \cellcolor{custom_green!20}4.5  & \cellcolor{custom_green!20}33.2 & \multicolumn{2}{c}{\textit{OOM}}  & 0.2  & 66.7 & \cellcolor{custom_green!70}5.6  & \cellcolor{custom_green!70}27.9 \\
\bottomrule
\end{tabular}
}
\begin{tablenotes}\scriptsize
\item[$\dag$]: Methods that do \emph{not} produce temporally consistent geometry.
\end{tablenotes}
\end{threeparttable}
\vspace{-6mm}
\label{tab:runtime}
\end{table}

\subsection{Ablation Study}
\label{sec:ablation}

\myheading{Ablation on Adapter.}
We investigate strategies to fuse \emph{multi-view–consistent} LR tokens into \emph{high-resolution, fine-grained} HR tokens (Tab.~\ref{tab:ablation_adapter}).
\textbf{Setting A (post-align):} per-frame MoGe2~\cite{moge2} with \emph{post hoc} alignment to a multi-view–consistent pointmap from a visual-geometry model (e.g.,~\cite{pi3}), termed aligned MoGe2; this improves detail but leaves layering/stitching artifacts (see Supp.).
\textbf{Setting B (interp+SA):} LR tokens interpolated to the HR grid, concatenated with HR tokens, then fused via several HR self-attention layers.
\textbf{Setting C (all-CA):} adapter blocks use cross-attention only, with HR queries attending to LR keys/values at every block.
\textbf{Setting D (Alter-Adapter):} a cross/self-attention module inserted after each of the last 5 ViT blocks in the HR stream.
Overall, the proposed \(\text{CrossAttn}-\text{SelfAttn}\) adapter, inserted after the HR-stream ViT, consistently outperforms these variants, reducing artifacts and improving cross-view coherence.

\begin{table}[t]
\centering
\caption{\textbf{Ablation studies} (a) different adapter design, (b) effect of each component on the depth sharpness}
\vspace{-3mm}
\captionsetup[subtable]{justification=centering}
\scriptsize
\setlength{\tabcolsep}{1pt}
\begin{subtable}[t]{0.45\linewidth}
\centering
\begin{tabular}{lcc}
\toprule
Variant & Acc.$\downarrow$ & Comp.$\downarrow$ \\
\midrule
A: Aligned MoGe2      &  0.031     & 0.028      \\
B: Interp+SA       &  0.030     & 0.024      \\
C: All-CA       &  0.021    & 0.018      \\
D: Alter-Adapter     &  0.023     & 0.018         \\
\cdashline{1-3}
\textbf{Ours}    &  \textbf{0.018}     & \textbf{0.016}      \\
\bottomrule
\end{tabular}
\caption{Adapter design (NRGBD~\cite{nrgbd})}
\label{tab:ablation_adapter}
\end{subtable}
\begin{subtable}[t]{0.54\linewidth}
\centering
\begin{tabular}{lccc}
\toprule
Variant & F1$\uparrow$ & Rel$^{p}\!\downarrow$ & $\delta^{p}\!\uparrow$ \\
\midrule
Pi3~\cite{pi3}+AnyUp~\cite{anyup} & 0.09  & 24.5  & 67.8 \\
\cdashline{1-4}
\textbf{Ours}               & \textbf{0.34} & 21.5  & \textbf{75.6}  \\
$-$ W/o mono. prior    & 0.27 & 22.6 & 73.5   \\
$-$ W/o gradient loss  & 0.31  & 21.4  & 75.5  \\
$+$ With local loss    & 0.30  & \textbf{20.9}  & 75.1  \\
\bottomrule
\end{tabular}
\caption{Depth sharpness (Sintel~\cite{sintel})}
\label{tab:ablation_sharp}
\end{subtable}
\vspace{-4mm}
\end{table}

\begin{figure}
    \centering
    \includegraphics[width=1.0\linewidth]{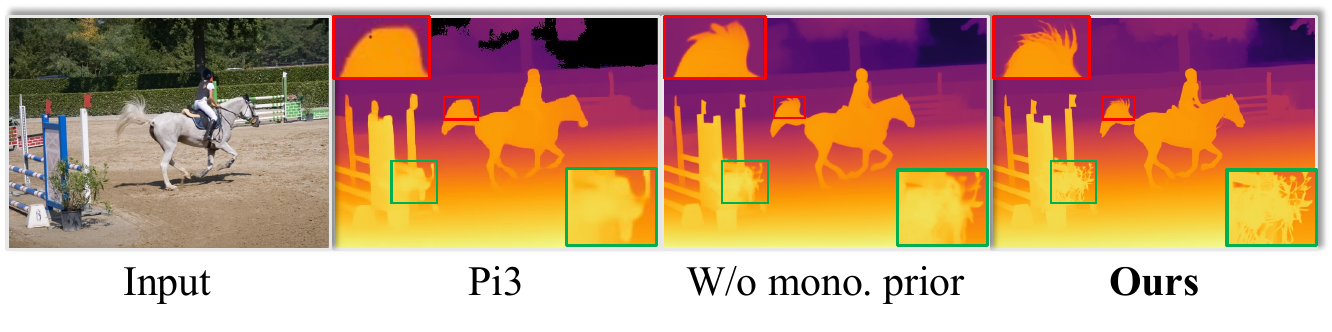}
    \vspace{-7mm}
    \caption{Comparison on disparity quality on Pi3, our variant without monocular prior, and our full model.}
    \vspace{-6mm} 
    \label{fig:ablation_sharp_depth}
\end{figure}

\myheading{Ablation on sharpness depth.}
Tab.~\ref{tab:ablation_sharp} ablates the contribution of the monocular prior~\cite{moge2} and the gradient loss; Fig.~\ref{fig:ablation_sharp_depth} shows qualitative results with and without the prior.

\myheading{Ablation on LR-stream resolution.} We vary the LR stream resolution in Tab.~\ref{tab:ablation_reso}. In general, increasing LR resolution slightly improves performance but significantly reduces the FPS.
\begin{table}
    \small
    \caption{\textbf{Ablation study} on the LR stream resolution}
    \vspace{-3mm}
    \setlength{\tabcolsep}{2pt}
    \centering
    \label{tab:ablation_reso}
    \resizebox{\linewidth}{!}{
    \begin{tabular}{l|ccc|ccc|cc|c}
    \toprule
    \multirow{2}{*}{Reso.} & \multicolumn{3}{c|}{\textbf{Sintel}~\cite{sintel} (pose)} & \multicolumn{3}{c|}{\textbf{Sintel}~\cite{sintel} (depth)} & \multicolumn{2}{c|}{\textbf{NRGBD}~\cite{nrgbd}}  & \multirow{2}{*}{FPS$\uparrow$} \\
     & ATE$\downarrow$ & RPE$_{T}$$\downarrow$ & RPE$_{R}$$\downarrow$ & Rel$^{p}\!\downarrow$ & $\delta^{p}\!\uparrow$ & F1$\uparrow$ & Acc.$\downarrow$ & Comp.$\downarrow$  \\
    \midrule
    252px (default)    & 0.132 & 0.051 & 0.406 & 21.5 & 75.5 & 0.34 & 0.018 & 0.016 & 65.4 \\
    \cdashline{1-10}
    252px (no distill) & 0.111 & 0.057 & 0.584 & 22.9 & 73.0 & 0.35 & 0.019 & 0.017 & 65.4 \\
    336px  & 0.130 & 0.042 & 0.307  & 20.4 & 76.9  & 0.35 & 0.018 & 0.016 & 46.6 \\
    518px  & 0.117  & 0.039 & 0.258  & 19.7 & 77.7 & 0.36 & 0.016 & 0.016 & 22.5 \\
    \bottomrule
    \end{tabular}
    \vspace{-10mm} 
    }
\end{table}

\section{Conclusion}
\label{sec:conclusion}

We introduced \Approach, a dual-stream visual geometry transformer. A low-resolution stream efficiently estimates cameras and enforces cross-view consistency, while a high-resolution stream preserves sharp details; a lightweight adapter fuses them. This decouples resolution from sequence length, supporting 2K inputs and long videos at practical costs. Empirically, \Approach yields sharper pointmaps and outperforms prior video geometry methods. It matches the 3D reconstruction and pose accuracy of state-of-the-art models~\cite{vggt,pi3} while running significantly faster. \textbf{Limitations.} Performance can drop under extremely low overlap or rapid non-rigid motion; the HR path is memory-intensive at very high resolutions; and the current method does not recover dynamic motion.

 \paragraph{Acknowledgements} Evangelos Kalogerakis has received funding from the European Research Council (ERC) under the Horizon research and innovation programme (Grant agreement No. 101124742).

\clearpage
\clearpage
{
    \small
    \bibliographystyle{ieeenat_fullname}
    \bibliography{main}
}

\clearpage
\setcounter{page}{1}
\maketitlesupplementary

\section{More Training Details}
\label{sec:supp_training_details}

\subsection{Training datasets}
\label{sec:supp_training_datasets}
We train on 18 datasets spanning indoor, outdoor, and object-centric scenes, covering both static and dynamic settings. The full list appears in Tab.~\ref{tab:train_datasets}. Following ~\cite{cut3r}, we filter scenes with ambiguous annotations in PointOdyssey~\cite{pointodyssey}, and remove scenes with panorama backgrounds and zoom-in/out effect in BEDLAM~\cite{bedlam}. For object-centric datasets~\cite{co3d,wildrgbd}, we only subsample 40 scenes for each object category.

\begin{table}[t]
\centering
\small
\caption{\textbf{Training datasets}.}
\setlength{\tabcolsep}{1pt}
\resizebox{\linewidth}{!}{
\begin{threeparttable}
    
\begin{tabular}{@{}lcccc@{}}
\toprule
\textbf{Dataset Name} & \textbf{Scene type} & \textbf{Metric} & \textbf{Real} & \textbf{Dynamic} \\
\midrule
ARKitScenes~\cite{arkitscenes}            & Indoor          & Yes & Real       & Static \\
ScanNet~\cite{scannet}                    & Indoor          & Yes & Real       & Static \\
ScanNet++~\cite{yeshwanth2023scannet++}                & Indoor          & Yes & Real       & Static \\
TartanAir~\cite{tartanair}                & Mixed           & Yes & Synthetic  & Dynamic \\
Waymo~\cite{waymo}                        & Outdoor         & Yes & Real       & Dynamic \\
BlendedMVS~\cite{blendedmvs}              & Mixed           & No  & Synthetic  & Static \\
HyperSim~\cite{hypersim}                  & Indoor          & Yes & Synthetic  & Static \\
MVS Synth~\cite{mvs_synth}                & Outdoor         & No  & Synthetic  & Static \\
GTA-Sfm~\cite{gta_sfm}                    & Outdoor         & No  & Synthetic  & Static \\
MegaDepth~\cite{megadepth}                & Outdoor         & No  & Real       & Static \\
CO3Dv2~\cite{co3d}$^\dagger$                     & Object-centric  & No  & Real       & Static \\
WildRGBD~\cite{wildrgbd}$^\dagger$                  & Object-centric  & Yes & Real       & Static \\
VirtualKITTI2~\cite{virtualkitti}        & Outdoor         & Yes & Synthetic  & Dynamic \\
Matterport3D~\cite{matterport3d}          & Indoor          & Yes & Real       & Static \\
BEDLAM~\cite{bedlam}$^\dagger$                      & Mixed           & Yes & Synthetic  & Dynamic \\
Dynamic~Replica~\cite{dynamicstereo}     & Indoor          & Yes & Synthetic  & Dynamic \\
PointOdyssey~\cite{pointodyssey}$^\dagger$          & Mixed           & Yes & Synthetic  & Dynamic \\
Spring~\cite{spring}                      & Mixed           & Yes & Synthetic  & Dynamic \\
\bottomrule
\end{tabular}
\begin{tablenotes}\small
\item[$\dagger$] \textit{Only a subset of each dataset is used.}
\end{tablenotes}
\end{threeparttable}
}
\label{tab:train_datasets}
\end{table}

\subsection{Implementation Details}

\myheading{Architecture.} For the high-resolution (HR) stream, we initialize the model with the 24-layer ViT from MoGe2~\cite{moge2} and keep these weights frozen throughout training. For the low-resolution (LR) stream, our training corpus (Sec.~\ref{sec:supp_training_datasets}) is considerably smaller than those used by recent feed-forward visual-geometry models~\cite{vggt,pi3,mapanything}. Consequently, rather than training a global video transformer from scratch, we start from the Pi3~\cite{pi3} checkpoint, which comprises 36 attention layers with alternating frame-wise and global attention. The adapter contains five blocks; each block consists of one cross-attention layer, one self-attention layer, and an MLP. We train the adapter from scratch and \emph{zero-initialize} its final projection to avoid destabilizing the frozen HR features at the start of training. For dense geometry, the pointmap head is implemented as a stack of residual convolutional blocks with transposed convolutions that progressively upsample from patch resolution \( (h_{hr} \times w_{hr} = H/14 \times W/14)\) to the original image resolution \((H \times W)\). Camera poses and the scene-wise metric scale factor are predicted with a two-layer MLP. For distillation during training, we use a 2-layer MLP and \textit{pixel shuffle} to project $\mathcal{F}^{\mathrm{lr}}$ to the teacher spatial resolution.

\myheading{Training loss.} We set the weight for each loss as follow: $\lambda_{\mathrm{pm}}=1.0, \lambda_{\mathrm{cam}}=0.1$, $\lambda_{\mathrm{trans}}=100.0$, $\lambda_{\mathrm{rot}}=1.0$, $\lambda_{\mathrm{scale}}=1.0$, $\lambda_{\mathrm{normal}}=1.0$, $\lambda_{\mathrm{gradient}}=0.1$, $\lambda_{\mathrm{distill}}=0.5$.

\myheading{Optimization.}
We train our model in two stages. \emph{Stage 1} targets low–medium resolutions with longer clips; \emph{Stage 2} fine-tunes on high-resolution inputs with short clips.
Both stages use AdamW~\cite{adamw} optimizer with a OneCycleLR schedule.
In Stage~1 (30{,}000 steps), we set a base LR of $1\times10^{-4}$ for the adapter and dense heads, and $1\times10^{-5}$ (10$\times$ lower) for the global transformer initialized from Pi3.
In Stage~2 (10{,}000 steps), we freeze the global transformer and fine-tune only the adapter and heads at $1\times10^{-5}$.
To keep training efficient, we use FlashAttention~\cite{dao2022flashattention,dao2023flashattention}, \texttt{bfloat16} mixed precision, gradient checkpointing, and gradient accumulation.
With this setup, training takes roughly five days on 16$\times$A100-80GB GPUs.

\myheading{Augmentation and sampling.}
We extend the MoGe~\cite{moge} augmentation pipeline to the multi-view setting and adopt stage-specific regimes.
\emph{Stage 1 (long sequence):} we sample 2–24 frames per clip and constrain the total pixels to $[1.0\times10^{5},\,2.55\times10^{5}]$, thereby enabling a large per-GPU batch of 48 images; we apply distillation only in this stage.
\emph{Stage 2 (high resolution):} we sample just 2–4 frames per clip, set the total pixels to $[2.7\times10^{5},\,9.0\times10^{5}]$ (roughly $518{\times}518$–$952{\times}952$ for 14-px patches), vary the aspect ratio within $[0.5,\,2.0]$, and use 24 images per GPU.


\section{More Evaluation Details}
This section details the datasets and metrics used in our experiments.


\subsection{Video geometry estimation.}
\myheading{Datasets.}
Following GeometryCrafter~\cite{geometrycrafter}, we configure each test dataset as follows:
\begin{itemize}[leftmargin=*]
\item \textbf{GMU Kitchens}~\cite{gmu_kitchen}: We use all scenarios, extract 110 frames per sequence with a stride of 2, and downsample the 1920p videos and depth maps to \(960\times512\).
\item \textbf{Monkaa}~\cite{monkaa}: We select 9 scenes and truncate each sequence to 110 frames at the native resolution of \(960\times512\).
\item \textbf{Sintel}~\cite{sintel}: We use all training sequences (21–50 frames) and crop from \(1024\times436\) to \(896\times448\).
\item \textbf{ScanNet}~\cite{scannet}: We evaluate 100 test scenes with 90 frames per video (stride 3), and center-crop each frame to \(640\times512\).
\item \textbf{KITTI}~\cite{kitti}: We use all sequences from the depth-annotated validation split; for longer videos we keep the first 110 frames (yielding 13 videos with 67–110 frames), and center-crop to \(768\times384\).
\item \textbf{Diode}~\cite{diode}: We use all 771 validation images at the default resolution of \(1024\times768\).
\end{itemize}

\noindent
In addition, we prepare two high-resolution evaluation sets:
\begin{itemize}[leftmargin=*]
\item \textbf{UrbanSyn}~\cite{urbansyn}: We sample ten clips of 100 frames each from the original 7000-frame sequences and keep the resolution at \(2048\times1024\).
\item \textbf{Unreal4K}~\cite{unreal4k}: We use all nine scenes, keep the first 100 frames per scene, and downsample to \(1920\times1080\).
\end{itemize}

\myheading{Metrics.} For the pointmap estimation, we report the mean relative point error $\mathrm{Rel}^{p}\!\downarrow=\|\hat{\mathbf{p}}-\mathbf{p}\|_{2}/\|\mathbf{p}\|_{2}$ and the inlier ratio $\delta^{p}\!\uparrow$, where a point is an inlier if $\|\hat{\mathbf{p}}-\mathbf{p}\|_{2}/\min(\|\mathbf{p}\|_{2},\|\hat{\mathbf{p}}\|_{2})<\tau$ (with $\tau{=}0.25$), averaged over valid pixels. Similarly, we leverage $\mathrm{Rel}^{d}\!\downarrow$ and $\delta^{d}\!\uparrow$ for depth estimation.

\begin{table*}[t]
\centering
\caption{\textbf{Scale-invariant video pointmap evaluation}. Results are aligned with the ground truth by optimizing a shared scale factor across the entire video.  We mark {\setlength{\fboxsep}{2pt}\colorbox{custom_green!70}{\strut best}} and {\setlength{\fboxsep}{2pt}\colorbox{custom_green!20}{\strut second-best}}.}
\small
\setlength{\tabcolsep}{1pt}
\renewcommand{\arraystretch}{1.15}
\resizebox{\textwidth}{!}{%
\begin{threeparttable}
\begin{tabular}{l|cc|cc|cc|cc|cc|cc|cc|cc|c}
\toprule
\multirow{3}{*}{Method} &
\multicolumn{2}{c}{\textbf{GMU}~\cite{gmu_kitchen}} &
\multicolumn{2}{c}{\textbf{Monkaa}~\cite{monkaa}} &
\multicolumn{2}{c}{\textbf{Sintel}~\cite{sintel}} &
\multicolumn{2}{c}{\textbf{ScanNet}~\cite{scannet}} &
\multicolumn{2}{c}{\textbf{KITTI}~\cite{kitti}} &
\multicolumn{2}{c}{\textbf{UrbanSyn}~\cite{urbansyn}} &
\multicolumn{2}{c}{\textbf{Unreal4K}~\cite{unreal4k}} &
\multicolumn{2}{c}{\textbf{Diode}~\cite{diode}} &
\multirow{2}{*}{\textbf{Rank} $\downarrow$}
\\
& 
$\mathrm{Rel}^{p}\!\downarrow$ & $\delta^{p}\!\uparrow$ &
$\mathrm{Rel}^{p}\!\downarrow$ & $\delta^{p}\!\uparrow$ &
$\mathrm{Rel}^{p}\!\downarrow$ & $\delta^{p}\!\uparrow$ &
$\mathrm{Rel}^{p}\!\downarrow$ & $\delta^{p}\!\uparrow$ &
$\mathrm{Rel}^{p}\!\downarrow$ & $\delta^{p}\!\uparrow$ &
$\mathrm{Rel}^{p}\!\downarrow$ & $\delta^{p}\!\uparrow$ &
$\mathrm{Rel}^{p}\!\downarrow$ & $\delta^{p}\!\uparrow$ &
$\mathrm{Rel}^{p}\!\downarrow$ & $\delta^{p}\!\uparrow$ & \\
\midrule
DepthPro~\cite{depthpro}            & 
10.5 & 92.7 & 27.9 & 51.2 & 55.0 & 37.5 & 9.3 & 95.0 & 11.7 & 93.6 & 22.5 & 61.1 & 96.1 & 1.2 &  30.3 & 58.1 & 7.6 \\
MoGe~\cite{moge}                    & 
21.4 & 69.0 & 27.7 & 58.3 & 29.5 & 59.8 & 13.4 & 88.2 & 8.6 & 95.6 & 13.4 & 89.9 & 34.0 & 55.7 & 30.3 & 53.4 & 6.9 \\
MoGe2~\cite{moge2}                  & 
19.7 & 72.1 & 30.8 & 51.1 & 34.3 & 47.7 & 12.7 & 89.4 & 11.7 & 96.9 & 12.3 & 91.7 & 30.1 & 62.3 & 29.5 & 55.4 & 7.0\\
MoGe2~\cite{moge2}$^{\dagger}$      &
7.1 & \cellcolor{custom_green!70}94.6 & 25.8 & 60.2 & 33.1 & 52.1 & 7.8 & 97.5 & 10.5 & 98.4 & \cellcolor{custom_green!70}6.5 & \cellcolor{custom_green!70}97.2 & \cellcolor{custom_green!20}8.9 & 92.1 & 15.8 & 84.1 & 3.9 \\
CUT3R~\cite{cut3r}                  & 
8.2 & 93.6 & 34.9 & 45.9 & 42.9 & 35.8 & 6.5 & 98.0 & 16.0 & 88.1 & 57.9 & 14.0 & 17.5 & 78.3 & 17.2 & 81.6 & 6.9  \\
VGGT~\cite{vggt}                    & 
5.6 & 93.8 & 16.0 & 80.4 & \cellcolor{custom_green!20}26.7 & \cellcolor{custom_green!20}65.8 & \cellcolor{custom_green!20}3.1 & 99.0 & 8.4 & 97.3 & 18.5 & 75.0 & \cellcolor{custom_green!70}8.7 & \cellcolor{custom_green!70}96.5 & 13.6 & 80.2 & 3.6 \\
Pi3~\cite{pi3}                      & 
\cellcolor{custom_green!20}5.4 & 94.2 & \cellcolor{custom_green!20}12.6 & \cellcolor{custom_green!70}90.2 & 29.6 & 62.5 & \cellcolor{custom_green!70}2.4 & \cellcolor{custom_green!20}99.4 & 9.2 & 90.8 & 10.7 & 93.8 & 17.2 & 75.4 & \cellcolor{custom_green!70}9.0 & \cellcolor{custom_green!70}96.1 & \cellcolor{custom_green!20}3.1\\
GeoCrafter~\cite{geometrycrafter}   & 
8.4 & \cellcolor{custom_green!20}94.5 & 20.7 & 73.9 & 30.2 & 57.8 & 8.9 & 96.4 & \cellcolor{custom_green!70}6.4 & \cellcolor{custom_green!20}98.8 & 11.3 & 95.3 & 21.0 & 73.5 & 13.0 & 92.8 & 4.1 \\
\cdashline{1-18}
\textbf{\Approach (ours)} & \cellcolor{custom_green!70}5.0 & 94.2 & \cellcolor{custom_green!70}11.3 & \cellcolor{custom_green!20}88.1 & \cellcolor{custom_green!70}26.6 & \cellcolor{custom_green!70}66.2 & \cellcolor{custom_green!70}2.4 & \cellcolor{custom_green!70}99.5 & \cellcolor{custom_green!20}7.3 & \cellcolor{custom_green!70}99.0 & \cellcolor{custom_green!20}7.9 & \cellcolor{custom_green!20}96.6 & 9.2 & \cellcolor{custom_green!20}92.9 & \cellcolor{custom_green!20}10.0 & \cellcolor{custom_green!20}94.4 & \cellcolor{custom_green!70}\textbf{1.7}\\   
\bottomrule
\end{tabular}
\end{threeparttable}
} 
\label{tab:supp_quanti_video_pointmap_scale_inv}
\end{table*}

\begin{table*}[t]
\centering
\caption{\textbf{Affine-invariant video depthmap evaluation}. Results are aligned with the ground truth by optimizing a shared scale and shift factor across the entire video.  We mark {\setlength{\fboxsep}{2pt}\colorbox{custom_green!70}{\strut best}} and {\setlength{\fboxsep}{2pt}\colorbox{custom_green!20}{\strut second-best}}.}
\small
\setlength{\tabcolsep}{1pt}
\renewcommand{\arraystretch}{1.15}
\resizebox{\textwidth}{!}{%
\begin{threeparttable}
\begin{tabular}{l|cc|cc|cc|cc|cc|cc|cc|cc|c}
\toprule
\multirow{3}{*}{Method} &
\multicolumn{2}{c}{\textbf{GMU}~\cite{gmu_kitchen}} &
\multicolumn{2}{c}{\textbf{Monkaa}~\cite{monkaa}} &
\multicolumn{2}{c}{\textbf{Sintel}~\cite{sintel}} &
\multicolumn{2}{c}{\textbf{ScanNet}~\cite{scannet}} &
\multicolumn{2}{c}{\textbf{KITTI}~\cite{kitti}} &
\multicolumn{2}{c}{\textbf{UrbanSyn}~\cite{urbansyn}} &
\multicolumn{2}{c}{\textbf{Unreal4K}~\cite{unreal4k}} &
\multicolumn{2}{c}{\textbf{Diode}~\cite{diode}} &
\multirow{2}{*}{\textbf{Rank} $\downarrow$}
\\
& 
$\mathrm{Rel}^{d}\!\downarrow$ & $\delta^{d}\!\uparrow$ &
$\mathrm{Rel}^{d}\!\downarrow$ & $\delta^{d}\!\uparrow$ &
$\mathrm{Rel}^{d}\!\downarrow$ & $\delta^{d}\!\uparrow$ &
$\mathrm{Rel}^{d}\!\downarrow$ & $\delta^{d}\!\uparrow$ &
$\mathrm{Rel}^{d}\!\downarrow$ & $\delta^{d}\!\uparrow$ &
$\mathrm{Rel}^{d}\!\downarrow$ & $\delta^{d}\!\uparrow$ &
$\mathrm{Rel}^{d}\!\downarrow$ & $\delta^{d}\!\uparrow$ &
$\mathrm{Rel}^{d}\!\downarrow$ & $\delta^{d}\!\uparrow$ & \\
\midrule
DepthPro~\cite{depthpro}            & 
8.8 & 93.0 & 23.3 & 55.8 & 36.1 & 49.5 & 8.1 & 94.6 & 11.7 & 93.6 & 51.3 & 38.3 & 105.0 & 20.0 & 31.0 & 58.8 & 8.0 \\
MoGe~\cite{moge}                    & 
19.9 & 66.5 & 19.9 & 63.6 & 26.5 & 60.0 & 12.5 & 85.9 & 7.6 & 94.2 & 15.2 & 82.1 & 38.7 & 46.2 & 31.3 & 48.2 & 7.6 \\
MoGe2~\cite{moge2}                  & 
19.0 & 68.9 & 20.8 & 60.8 & 26.4 & 59.9 & 12.1 & 86.9 & 6.7 & 96.7 & 13.8 & 85.4 & 33.6 & 52.8 & 29.5 & 50.4 & 6.8 \\
MoGe2~\cite{moge2}$^{\dagger}$      &
6.6 & \cellcolor{custom_green!20}93.8 & 18.0 & 68.1 & 25.0 & 63.8 & 6.4 & 96.8 & 5.2 & 98.3 & \cellcolor{custom_green!70}7.7 & \cellcolor{custom_green!70}96.4 & \cellcolor{custom_green!20}12.0 & \cellcolor{custom_green!20}88.3 & 14.7 & 80.7 & 3.8 \\
CUT3R~\cite{cut3r}                  & 
7.3 & 92.9 & 28.0 & 49.9 & 31.9 & 50.5 & 5.4 & 97.4 & 10.2 & 89.1 & 48.6 & 36.7 & 15.4 & 79.7 & 16.0 & 78.4 & 6.8 \\
VGGT~\cite{vggt}                    & 
5.2 & 93.2 & 12.3 & 80.5 & 22.2 & 70.4 & 2.7 & 98.7 & 4.7 & 97.2 & 13.9 & 84.5 & \cellcolor{custom_green!70}8.2 & \cellcolor{custom_green!70}94.3 & 12.4 & 85.2 & 3.5  \\
Pi3~\cite{pi3}                      & 
\cellcolor{custom_green!20}4.9 & 93.5 & \cellcolor{custom_green!70}8.2 & \cellcolor{custom_green!70}91.4 & \cellcolor{custom_green!20}20.2 & \cellcolor{custom_green!20}71.7 & \cellcolor{custom_green!70}2.0 & \cellcolor{custom_green!20}99.3 & \cellcolor{custom_green!70}3.0 & \cellcolor{custom_green!70}99.1 & 16.0 & 78.8 & 18.3 & 78.6 & \cellcolor{custom_green!70}8.6 & \cellcolor{custom_green!20}92.9 & \cellcolor{custom_green!20}2.7 \\
GeoCrafter~\cite{geometrycrafter}   & 7.7 & \cellcolor{custom_green!70}94.1 & 13.4 & 79.3 & 21.4 & 70.6 & 7.3 & 96.1 & 5.0 & 98.5 & 12.2 & 90.3 & 20.7 & 72.2 & 9.1 & \cellcolor{custom_green!70}93.4 & 3.9 \\
\cdashline{1-18}
\textbf{\Approach (ours)}  & \cellcolor{custom_green!70}4.8 & 93.5 & \cellcolor{custom_green!20}9.5 & \cellcolor{custom_green!20}87.2 & \cellcolor{custom_green!70}19.5 & \cellcolor{custom_green!70}74.4 & \cellcolor{custom_green!20}2.1 & \cellcolor{custom_green!70}99.4 & \cellcolor{custom_green!20}3.2 & \cellcolor{custom_green!20}98.8 & \cellcolor{custom_green!20}7.7 & \cellcolor{custom_green!20}95.8 & 12.1 & 88.1 & \cellcolor{custom_green!20}8.7 & 92.5 & \cellcolor{custom_green!70}\textbf{1.9} \\   
\bottomrule
\end{tabular}
\end{threeparttable}
} 
\label{tab:supp_quanti_video_depth_affine_inv}
\end{table*}

\begin{table*}[t]
\centering
\caption{\textbf{Scale-invariant video depthmap evaluation}. Results are aligned with the ground truth by optimizing a shared scale factor across the entire video.  We mark {\setlength{\fboxsep}{2pt}\colorbox{custom_green!70}{\strut best}} and {\setlength{\fboxsep}{2pt}\colorbox{custom_green!20}{\strut second-best}}.}
\small
\setlength{\tabcolsep}{1pt}
\renewcommand{\arraystretch}{1.15}
\resizebox{\textwidth}{!}{%
\begin{threeparttable}
\begin{tabular}{l|cc|cc|cc|cc|cc|cc|cc|cc|c}
\toprule
\multirow{3}{*}{Method} &
\multicolumn{2}{c}{\textbf{GMU}~\cite{gmu_kitchen}} &
\multicolumn{2}{c}{\textbf{Monkaa}~\cite{monkaa}} &
\multicolumn{2}{c}{\textbf{Sintel}~\cite{sintel}} &
\multicolumn{2}{c}{\textbf{ScanNet}~\cite{scannet}} &
\multicolumn{2}{c}{\textbf{KITTI}~\cite{kitti}} &
\multicolumn{2}{c}{\textbf{UrbanSyn}~\cite{urbansyn}} &
\multicolumn{2}{c}{\textbf{Unreal4K}~\cite{unreal4k}} &
\multicolumn{2}{c}{\textbf{Diode}~\cite{diode}} &
\multirow{2}{*}{\textbf{Rank} $\downarrow$}
\\
& 
$\mathrm{Rel}^{d}\!\downarrow$ & $\delta^{d}\!\uparrow$ &
$\mathrm{Rel}^{d}\!\downarrow$ & $\delta^{d}\!\uparrow$ &
$\mathrm{Rel}^{d}\!\downarrow$ & $\delta^{d}\!\uparrow$ &
$\mathrm{Rel}^{d}\!\downarrow$ & $\delta^{d}\!\uparrow$ &
$\mathrm{Rel}^{d}\!\downarrow$ & $\delta^{d}\!\uparrow$ &
$\mathrm{Rel}^{d}\!\downarrow$ & $\delta^{d}\!\uparrow$ &
$\mathrm{Rel}^{d}\!\downarrow$ & $\delta^{d}\!\uparrow$ &
$\mathrm{Rel}^{d}\!\downarrow$ & $\delta^{d}\!\uparrow$ & \\
\midrule
DepthPro~\cite{depthpro}            & 
9.4 & 92.1 & 26.7 & 45.9 & 53.6 & 35.3 & 8.8 & 92.9 & 8.2 & 92.5 & 22.2 & 40.2 & 96.0 & 1.1 & 29.3 & 56.4 & 7.9 \\
MoGe~\cite{moge}                    & 
20.7 & 64.7 & 25.5 & 54.8 & 31.4 & 48.9 & 13.3 & 85.0 & 7.7 & 94.1 & 13.1 & 86.3 & 34.7 & 49.8 & 29.8 & 48.2 & 7.6 \\
MoGe2~\cite{moge2}                  & 
19.5 & 67.1 & 27.1 & 51.6 & 31.2 & 47.7 & 12.0 & 86.5 & 7.2 & 96.7 & 12.0 & 88.8 & 30.7 & 56.6 & 28.5 & 50.7 & 6.9 \\
MoGe2~\cite{moge2}$^{\dagger}$      &
6.7 & \cellcolor{custom_green!70}93.8 & 21.9 & 60.4 & 30.1 & 51.9 & 7.1 & 95.9 & 5.6 & 98.3 & \cellcolor{custom_green!70}6.0 & \cellcolor{custom_green!70}96.8 & \cellcolor{custom_green!20}8.7 & \cellcolor{custom_green!20}91.0 & 14.7 & 80.7 & 3.7 \\
CUT3R~\cite{cut3r}                  & 
7.9 & 92.6 & 33.0 & 38.3 & 37.3 & 42.4 & 5.8 & 97.0 & 11.3 & 86.8 & 22.2 & 63.1 & 16.8 & 79.6 & 15.6 & 80.0 & 6.7 \\
VGGT~\cite{vggt}                    & 
5.2 & 93.0 & 14.4 & 77.3 & \cellcolor{custom_green!70}25.3 & \cellcolor{custom_green!20}62.1 & 2.8 & 98.6 & 5.3 & 96.7 & 18.3 & 73.3 & \cellcolor{custom_green!70}8.2 & \cellcolor{custom_green!70}96.1 & 13.4 & 79.2 & 3.6 \\
Pi3~\cite{pi3}                      & 
\cellcolor{custom_green!20}4.9 & 93.4 & \cellcolor{custom_green!70}10.8 & \cellcolor{custom_green!70}88.9 & 28.4 & 60.6 & \cellcolor{custom_green!70}2.1 & \cellcolor{custom_green!20}99.3 & \cellcolor{custom_green!70}3.1 & \cellcolor{custom_green!70}99.1 & 9.5 & 92.5 & 16.6 & 75.0 & \cellcolor{custom_green!70}8.7 & \cellcolor{custom_green!70}95.5 & \cellcolor{custom_green!20}2.3 \\
GeoCrafter~\cite{geometrycrafter}   & 
8.1 & \cellcolor{custom_green!20}93.8 & 18.1 & 71.1 & 27.1 & 58.7 & 7.9 & 95.5 & 5.1 & 98.4 & 11.0 & 92.4 & 21.1 & 70.9 & 10.0 & 92.4 & 4.1 \\
\cdashline{1-18}
\textbf{\Approach (ours)} & \cellcolor{custom_green!70}4.7 & 93.4 & \cellcolor{custom_green!20}11.5 & \cellcolor{custom_green!20}85.5 & \cellcolor{custom_green!20}25.6 & \cellcolor{custom_green!70}64.8 & \cellcolor{custom_green!20}2.2 & \cellcolor{custom_green!70}99.4 & \cellcolor{custom_green!20}3.3 & \cellcolor{custom_green!20}98.8 & \cellcolor{custom_green!20}7.9 & \cellcolor{custom_green!20}95.9 & 8.7 & 90.3 & \cellcolor{custom_green!20}9.9 & \cellcolor{custom_green!20}94.0 & \cellcolor{custom_green!70}\textbf{1.9}
\\   
\bottomrule
\end{tabular}
\end{threeparttable}
} 
\label{tab:supp_quanti_video_depth_scale_inv}
\end{table*}

\subsection{Video sharpness depth.}
\myheading{Datasets.} We evaluate depth–boundary sharpness on four synthetic datasets—Monkaa~\cite{monkaa}, Sintel~\cite{sintel}, UrbanSyn~\cite{urbansyn}, and Unreal4K~\cite{unreal4k}.

\myheading{Metrics.}
We use the F1$\!\uparrow$ edge metric from DepthPro~\cite{depthpro}. For each pair of neighboring pixels, we mark an occluding contour when the depth ratio exceeds a \textit{predefined threshold}. Applying this to both prediction and ground truth yields two contour maps. \textit{Precision} is the fraction of predicted contour pairs that are also contours in the ground truth, and \textit{recall} is the fraction of ground-truth contour pairs recovered by the prediction. The F1 score is the harmonic mean of precision and recall. We report the F1 averaged over multiple thresholds. This metric requires no ground-truth edge maps and is easily computed wherever dense depth annotations are available (e.g., synthetic data).

\noindent
To further assess boundary sharpness, we adopt the Depth Boundary Error (DBE) from iBims~\cite{ibims} and use its pseudo variant (PDBE) for datasets without depth–edge annotations (following~\cite{sharpdepth}). Concretely, we run Canny edge detection on both predicted and ground-truth depth maps to obtain edge sets, then compute the iBims accuracy and completeness terms. The accuracy term penalizes predicted edges that are far from any ground-truth edge, while the completeness term penalizes ground-truth edges not recovered by the prediction. Finally, we report the \textit{chamfer distance} $\mathcal{C}_{\mathrm{PDBE}}\!\downarrow$, which is the average of accuracy and completeness.

\subsection{Multi-view reconstruction.} 
\myheading{Datasets.} We evaluate 3D pointmap reconstruction on 7-Scenes~\cite{7scenes} and NRGBD~\cite{nrgbd} under both sparse and dense view protocols. For sparse views, we sample keyframes every 200 frames on 7-Scenes and every 500 on NRGBD; for dense views, the strides are 40 and 100, respectively.

\myheading{Metrics.}
We employ the \emph{Accuracy} (Acc$\downarrow$): mean nearest-neighbor distance from each predicted point to the ground truth,
\emph{Completion} (Comp$\downarrow$): mean nearest-neighbor distance from each ground-truth point to the reconstruction,
and \emph{Normal Consistency} (NC$\uparrow$): mean absolute dot product of ground truth and predicted normals (computed on the fly using \texttt{Open3D} library). 



\subsection{Camera pose estimation.}
\myheading{Datasets.} We evaluate on Sintel~\cite{sintel}, TUM-Dynamics~\cite{tum_dynamic}, and ScanNet~\cite{scannet}. For Sintel, we follow ~\cite{leapvo,monst3r}, excluding static scenes and those with perfectly straight camera motion, leaving 14 sequences. For TUM-Dynamics and ScanNet, we use the first 90 frames with a temporal stride of 3. 

\myheading{Metrics.}
Following \cite{monst3r,pi3,cut3r}, we report Absolute Trajectory Error (ATE$\!\downarrow$) and Relative Pose Error for translation and rotation (RPE$_T\!\downarrow$ / RPE$_R\!\downarrow$). Predicted trajectories are first aligned to ground truth with a single $\mathrm{Sim}(3)$ transform (global scale, rotation, translation). ATE is the root-mean-square discrepancy between aligned and ground-truth camera positions over the entire sequence.
RPE$_T$ is the translation error over a certain distance, and RPE$_R$ is the rotation error over a certain degree; both are averaged over all pose pairs.

\section{More Results}
\subsection{Video geometry estimation}
We evaluate video geometry estimation under four other settings. First, for \emph{scale-invariant} video pointmaps, we align predictions to ground truth with a \emph{single} per-video scale and report results in Tab.~\ref{tab:supp_quanti_video_pointmap_scale_inv}. Second, for video \emph{depth}, we follow standard practice and report both \emph{affine-invariant} results—per-frame scale\,+\,shift alignment—in Tab.~\ref{tab:supp_quanti_video_depth_affine_inv}, and \emph{scale-invariant} results—single per-video scale—in Tab.~\ref{tab:supp_quanti_video_depth_scale_inv}. Finally, we assess \emph{metric-scale} video pointmaps with \textbf{no} alignment (direct comparison in the dataset’s metric units); see Tab.~\ref{tab:supp_quanti_video_pointmap_metric}. For the metric setting, we compare against methods capable of predicting metric geometry, including CUT3R~\cite{cut3r} and MapAnything~\cite{mapanything}.

We additionally evaluate feed-forward visual-geometry approaches at each dataset’s native resolution (540p–2K). As reported in Tab.~\ref{tab:supp_quanti_video_point_affine_inv_hr}, performance degrades steadily with increasing resolution; at the highest, far beyond training scales (e.g. Urbansyn and Unreak4k datasets), most methods collapse except ours.

\begin{table*}[t]
\centering
\caption{\textbf{Metric video pointmap evaluation}. Predicted pointmaps are directly compared with ground truth.}
\small
\setlength{\tabcolsep}{3pt}
\renewcommand{\arraystretch}{1.15}
\begin{threeparttable}
\begin{tabular}{l|cc|cc|cc|cc|cc}
\toprule
\multirow{3}{*}{Method} &
\multicolumn{2}{c}{\textbf{GMU}~\cite{gmu_kitchen}} &
\multicolumn{2}{c}{\textbf{ScanNet}~\cite{scannet}} &
\multicolumn{2}{c}{\textbf{KITTI}~\cite{kitti}} &
\multicolumn{2}{c}{\textbf{UrbanSyn}~\cite{urbansyn}} &
\multicolumn{2}{c}{\textbf{Diode}~\cite{diode}} 
\\
& 
$\mathrm{Rel}^{p}\!\downarrow$ & $\delta^{p}\!\uparrow$ &
$\mathrm{Rel}^{p}\!\downarrow$ & $\delta^{p}\!\uparrow$ &
$\mathrm{Rel}^{p}\!\downarrow$ & $\delta^{p}\!\uparrow$ &
$\mathrm{Rel}^{p}\!\downarrow$ & $\delta^{p}\!\uparrow$ &
$\mathrm{Rel}^{p}\!\downarrow$ & $\delta^{p}\!\uparrow$ \\
\midrule
CUT3R~\cite{cut3r}                  & \cellcolor{custom_green!20}13.5 & \cellcolor{custom_green!20}90.7 & \cellcolor{custom_green!20}9.1 & \cellcolor{custom_green!20}95.2 & 34.2 & 14.6 & \cellcolor{custom_green!20}15.4 & \cellcolor{custom_green!20}84.4 & \cellcolor{custom_green!20}31.6 & \cellcolor{custom_green!20}47.2 
\\
MapAny~\cite{mapanything}                  & 22.6 & 63.4 & 35.2 & 28.5 & \cellcolor{custom_green!20}29.1 & \cellcolor{custom_green!20}26.3 & 28.5 & 37.3 & 33.8 & 31.8
\\
\cdashline{1-11}
\textbf{\Approach (ours)}                        & \cellcolor{custom_green!70}7.5 & \cellcolor{custom_green!70}95.3 & \cellcolor{custom_green!70}2.5 & \cellcolor{custom_green!70}99.5 & \cellcolor{custom_green!70}12.0 & \cellcolor{custom_green!70}98.3 & \cellcolor{custom_green!70}8.3 & \cellcolor{custom_green!70}96.5 & \cellcolor{custom_green!70}12.9 & \cellcolor{custom_green!70}87.5
 \\   
\bottomrule
\end{tabular}
\end{threeparttable}
\label{tab:supp_quanti_video_pointmap_metric}
\end{table*}

\begin{table*}[t]
\centering
\caption{\textbf{Affine-invariant video pointmap evaluation at native resolution.}
Predictions are aligned to ground truth by optimizing a single scale and shift across the entire video.}
\small
\setlength{\tabcolsep}{1pt}
\renewcommand{\arraystretch}{1.15}
\resizebox{\textwidth}{!}{%
\begin{threeparttable}
\begin{tabular}{l|cc|cc|cc|cc|cc|cc|cc|cc}
\toprule
\multirow{3}{*}{Method} &
\multicolumn{2}{c}{\textbf{GMU}~\cite{gmu_kitchen}} &
\multicolumn{2}{c}{\textbf{Monkaa}~\cite{monkaa}} &
\multicolumn{2}{c}{\textbf{Sintel}~\cite{sintel}} &
\multicolumn{2}{c}{\textbf{ScanNet}~\cite{scannet}} &
\multicolumn{2}{c}{\textbf{KITTI}~\cite{kitti}} &
\multicolumn{2}{c}{\textbf{UrbanSyn}~\cite{urbansyn}} &
\multicolumn{2}{c}{\textbf{Unreal4K}~\cite{unreal4k}} &
\multicolumn{2}{c}{\textbf{Diode}~\cite{diode}} \\
& \multicolumn{2}{c|}{($960\times512$)} & \multicolumn{2}{c|}{($960\times512$)} & \multicolumn{2}{c|}{($896\times448$)} & \multicolumn{2}{c|}{($640\times512$)} & \multicolumn{2}{c|}{($768\times384$)} & \multicolumn{2}{c|}{($2048\times1024$)} & \multicolumn{2}{c|}{($1920\times1080$)} & \multicolumn{2}{c|}{($1024\times768$)} \\
& 
$\mathrm{Rel}^{d}\!\downarrow$ & $\delta^{d}\!\uparrow$ &
$\mathrm{Rel}^{d}\!\downarrow$ & $\delta^{d}\!\uparrow$ &
$\mathrm{Rel}^{d}\!\downarrow$ & $\delta^{d}\!\uparrow$ &
$\mathrm{Rel}^{d}\!\downarrow$ & $\delta^{d}\!\uparrow$ &
$\mathrm{Rel}^{d}\!\downarrow$ & $\delta^{d}\!\uparrow$ &
$\mathrm{Rel}^{d}\!\downarrow$ & $\delta^{d}\!\uparrow$ &
$\mathrm{Rel}^{d}\!\downarrow$ & $\delta^{d}\!\uparrow$ &
$\mathrm{Rel}^{d}\!\downarrow$ & $\delta^{d}\!\uparrow$ \\
\midrule
CUT3R~\cite{cut3r}                  & 22.0 & 67.9 & 30.9 & 51.0 & 38.9 & 40.9 & 7.0 & 97.9 & 13.2 & 88.7 & 56.7 & 13.9 & 71.6 & 5.6 & 31.1 & 52.6
\\
VGGT~\cite{vggt}                    & 15.9 & 91.4 & 17.7 & 81.6 & 28.7 & 63.8 & 4.5 & 99.1 & 7.8 & 97.5 & OOM & OOM & OOM & OOM & 20.5 & 76.3 
\\
Pi3~\cite{pi3}                      & 6.2 & 92.2 & 12.6 & 88.9 & 21.7 & 72.9 & 2.2 & 99.5 & 5.9 & 97.5 & 55.9 & 14.7 & 54.2 & 17.1 & 13.9 & 87.2
 \\
\cdashline{1-17}
\textbf{\Approach (ours)} & \cellcolor{custom_green!70}4.9 & \cellcolor{custom_green!70}94.2 &
\cellcolor{custom_green!70}10.1 & \cellcolor{custom_green!70}91.0 &
\cellcolor{custom_green!70}21.5 & \cellcolor{custom_green!70}75.6 &
\cellcolor{custom_green!70}2.1 & \cellcolor{custom_green!70}99.5 &
\cellcolor{custom_green!70}5.9 & \cellcolor{custom_green!70}99.0 &
\cellcolor{custom_green!70}8.8 & \cellcolor{custom_green!70}96.0 &
\cellcolor{custom_green!70}11.9 & \cellcolor{custom_green!70}89.1 &
\cellcolor{custom_green!70}9.7 & \cellcolor{custom_green!70}94.4
\\
\bottomrule
\end{tabular}
\end{threeparttable}
} 
\label{tab:supp_quanti_video_point_affine_inv_hr}
\end{table*}



\subsection{Single-image geometry estimation}
Following \cite{moge,moge2}, we evaluate the single-image geometry estimation on eight different datasets, including NYUv2~\cite{NYUv2}, KITTI~\cite{kitti}, ETH3D~\cite{eth3d}, iBims-1~\cite{ibims}, GSO~\cite{GSO}, Sintel~\cite{sintel}, DDAD~\cite{ddad}, DIODE~\cite{diode}, HAMMER~\cite{HAMMER}. The results are summaried in Tab.~\ref{tab:supp_quanti_single_image_geometry}, validating that our dual-stream design preserves single-image quality compared to single-image based methods like DepthPro~\cite{depthpro}, MoGE~\cite{moge,moge2}.

\begin{table*}[t]
\centering
\caption{\textbf{Single-image geometry evaluation}. Results are aligned with the ground truth by optimizing a scale and shift factor for each image.  We mark {\setlength{\fboxsep}{2pt}\colorbox{custom_green!70}{\strut best}} and {\setlength{\fboxsep}{2pt}\colorbox{custom_green!20}{\strut second-best}}.}
\small
\setlength{\tabcolsep}{1pt}
\renewcommand{\arraystretch}{1.15}
\resizebox{\textwidth}{!}{%
\begin{threeparttable}
\begin{tabular}{l|cc|cc|cc|cc|cc|cc|cc|cc|cc|c}
\toprule
\multirow{3}{*}{Method} &
\multicolumn{2}{c}{\textbf{NYUv2}~\cite{NYUv2}} &
\multicolumn{2}{c}{\textbf{KITTI}~\cite{kitti}} &
\multicolumn{2}{c}{\textbf{ETH3D}~\cite{eth3d}} &
\multicolumn{2}{c}{\textbf{iBims-1}~\cite{ibims}} &
\multicolumn{2}{c}{\textbf{GSO}~\cite{GSO}} &
\multicolumn{2}{c}{\textbf{Sintel}~\cite{sintel}} &
\multicolumn{2}{c}{\textbf{DDAD}~\cite{ddad}} &
\multicolumn{2}{c}{\textbf{DIODE}~\cite{diode}} &
\multicolumn{2}{c}{\textbf{HAMMER}~\cite{HAMMER}} &
\multirow{2}{*}{\textbf{Rank} $\downarrow$}
\\
& 
$\mathrm{Rel}^{d}\!\downarrow$ & $\delta^{d}\!\uparrow$ &
$\mathrm{Rel}^{d}\!\downarrow$ & $\delta^{d}\!\uparrow$ &
$\mathrm{Rel}^{d}\!\downarrow$ & $\delta^{d}\!\uparrow$ &
$\mathrm{Rel}^{d}\!\downarrow$ & $\delta^{d}\!\uparrow$ &
$\mathrm{Rel}^{d}\!\downarrow$ & $\delta^{d}\!\uparrow$ &
$\mathrm{Rel}^{d}\!\downarrow$ & $\delta^{d}\!\uparrow$ &
$\mathrm{Rel}^{d}\!\downarrow$ & $\delta^{d}\!\uparrow$ &
$\mathrm{Rel}^{d}\!\downarrow$ & $\delta^{d}\!\uparrow$ &
$\mathrm{Rel}^{d}\!\downarrow$ & $\delta^{d}\!\uparrow$ \\
\midrule
DepthPro~\cite{depthpro}            & 
4.36 & 97.9 & 9.15 & 90.7 & 7.73 & 94.0 & 4.34 & 97.4 & 3.16 & 99.7 & 19.6 & 74.5 & 14.4 & 81.2 & 6.28 & 93.7 & 5.31 & \cellcolor{custom_green!20}98.8 & 3.8 \\
MoGe~\cite{moge}                    & 
3.68 & \cellcolor{custom_green!20}98.3 & \cellcolor{custom_green!20}4.86 & \cellcolor{custom_green!70}97.2 & \cellcolor{custom_green!20}3.57 & \cellcolor{custom_green!70}99.0 & \cellcolor{custom_green!20}3.61 & 97.3 & \cellcolor{custom_green!20}1.14 & \cellcolor{custom_green!70}100 & \cellcolor{custom_green!20}16.8 & \cellcolor{custom_green!70}77.8 & \cellcolor{custom_green!20}10.5 & \cellcolor{custom_green!70}91.4 & \cellcolor{custom_green!70}4.37 & \cellcolor{custom_green!70}96.4 & \cellcolor{custom_green!20}3.88 & 98.1 & \cellcolor{custom_green!70}1.8 \\
MoGe2~\cite{moge2}                  & 
\cellcolor{custom_green!70}3.33 & \cellcolor{custom_green!70}98.4 & \cellcolor{custom_green!70}6.47 & \cellcolor{custom_green!20}96.4 & 3.89 & \cellcolor{custom_green!20}98.7 & \cellcolor{custom_green!70}3.65 & \cellcolor{custom_green!70}98.5 & \cellcolor{custom_green!70}1.16 & \cellcolor{custom_green!70}100 & \cellcolor{custom_green!70}17.4 & \cellcolor{custom_green!20}77.0 & \cellcolor{custom_green!70}10.1 & \cellcolor{custom_green!20}90.3 & 5.13 & \cellcolor{custom_green!20}94.9 & 4.19 & \cellcolor{custom_green!70}99.1 & \cellcolor{custom_green!20}1.9 \\
\textbf{\Approach (ours)}  & \cellcolor{custom_green!20}3.34 & \cellcolor{custom_green!70}98.4 & 7.52 & 94.7 & \cellcolor{custom_green!70}3.49 & 98.0 & 3.70 & \cellcolor{custom_green!20}97.8 & 1.26 & \cellcolor{custom_green!20}99.9 & 18.9 & 74.8 & 10.7 & 89.2 & \cellcolor{custom_green!20}4.97 & 94.6 & \cellcolor{custom_green!70}3.43 & 98.6 & 2.5 \\ 
\bottomrule
\end{tabular}
\end{threeparttable}
} 
\label{tab:supp_quanti_single_image_geometry}
\end{table*}

\subsection{Camera pose estimation}
We additionally report the predicted camera poses on RealEstate10K and CO3Dv2 datasets. We report the Relative Rotation Accuracy (RRA) and Relative Translation Accuracy (RTA) at a given threshold, and the Area Under the Curve (AUC) of the min(RRA,RTA) threshold curve. Tab.~\ref{tab:supp_pose_re10k_co3dv2} shows that \Approach remains competitive with Pi3~\cite{pi3} and VGGT~\cite{vggt}, even while operating at a lower resolution.

\begin{table*}[t]
\centering
\caption{\textbf{Pose Estimation} on RealEstate10K and Co3Dv2}
\label{tab:supp_pose_re10k_co3dv2}
\small
\setlength{\tabcolsep}{4pt}
\begin{tabular}{lcccccc}
\toprule
\multirow{2}{*}{\textbf{Method}} & \multicolumn{3}{c}{\textbf{RealEstate10K}} & \multicolumn{3}{c}{\textbf{Co3Dv2}} \\
\cmidrule(lr){2-4} \cmidrule(lr){5-7}
& RRA@30 $\uparrow$ & RTA@30 $\uparrow$ & AUC@30 $\uparrow$ & RRA@30 $\uparrow$ & RTA@30 $\uparrow$ & AUC@30 $\uparrow$ \\
\midrule
VGGT (518px) & 99.97 & 93.13 & 77.62 & \cellcolor{custom_green!20}98.64 & \cellcolor{custom_green!20}97.62 & \cellcolor{custom_green!20}91.28 \\
Pi3 (518px) & \cellcolor{custom_green!70}99.99 & \cellcolor{custom_green!70}95.62 & \cellcolor{custom_green!70}85.90 & 98.49 & 97.53 & \cellcolor{custom_green!70}91.39 \\
\textbf{\Approach} (252px) & \cellcolor{custom_green!20}99.98 &  \cellcolor{custom_green!20}95.22 & \cellcolor{custom_green!20}83.12 & \cellcolor{custom_green!70}98.74 & \cellcolor{custom_green!70}97.71 & 90.71 \\
\bottomrule
\end{tabular}
\end{table*}

\subsection{More ablation studies}

\myheading{Low-resolution stream architecture.}
We perform an ablation study of the global module in our LR stream. Specifically, in addition to the global transformer with alternative frame/global attention, we ablate with two other design: (1) transformer-based recurrent network~\cite{cut3r} and (2) temporal Mamba network~\cite{flashdepth}. Results in Tab.~\ref{tab:supp_quanti_lr_stream_arch} show that the alternating global-attention transformer consistently outperforms both variants, reflecting stronger multi-view aggregation and more reliable cross-view consistency.

\myheading{RoPE design in the adapter.}
We ablate rotary positional encodings (RoPE) in the adapter in Tabs.~\ref{tab:supp_abl_rope_selfattn} and ~\ref{tab:supp_abl_rope_crossattn}. 
For self-attention (Tab.~\ref{tab:supp_abl_rope_selfattn}), standard RoPE~\cite{rope} is ineffective at high resolutions (e.g., UrbanSyn dataset), whereas interpolated RoPE improves performance. For cross-attention (Tab.~\ref{tab:supp_abl_rope_crossattn}), adding RoPE alongside our alignment (“snapping”) further boosts results.

\begin{table*}[!t]  
\centering
\captionsetup[subtable]{justification=centering}
\caption{\textbf{Ablations.} (a) LR-stream architectures. (b,c) Positional encodings.}
\vspace{-2mm}

\begin{subtable}{\textwidth}
\centering
\caption{\textbf{Ablation on different architectures of the LR stream}.}
\small
\setlength{\tabcolsep}{1pt}
\renewcommand{\arraystretch}{1.15}
\resizebox{0.9\textwidth}{!}{%
\begin{threeparttable}
\begin{tabular}{l|cc|cc|cc|cc|cc|cc|cc|cc}
\toprule
\multirow{3}{*}{Method} &
\multicolumn{2}{c}{\textbf{GMU}~\cite{gmu_kitchen}} &
\multicolumn{2}{c}{\textbf{Monkaa}~\cite{monkaa}} &
\multicolumn{2}{c}{\textbf{Sintel}~\cite{sintel}} &
\multicolumn{2}{c}{\textbf{ScanNet}~\cite{scannet}} &
\multicolumn{2}{c}{\textbf{KITTI}~\cite{kitti}} &
\multicolumn{2}{c}{\textbf{UrbanSyn}~\cite{urbansyn}} &
\multicolumn{2}{c}{\textbf{Unreal4K}~\cite{unreal4k}} &
\multicolumn{2}{c}{\textbf{Diode}~\cite{diode}} \\
& 
$\mathrm{Rel}^{p}\!\downarrow$ & $\delta^{p}\!\uparrow$ &
$\mathrm{Rel}^{p}\!\downarrow$ & $\delta^{p}\!\uparrow$ &
$\mathrm{Rel}^{p}\!\downarrow$ & $\delta^{p}\!\uparrow$ &
$\mathrm{Rel}^{p}\!\downarrow$ & $\delta^{p}\!\uparrow$ &
$\mathrm{Rel}^{p}\!\downarrow$ & $\delta^{p}\!\uparrow$ &
$\mathrm{Rel}^{p}\!\downarrow$ & $\delta^{p}\!\uparrow$ &
$\mathrm{Rel}^{p}\!\downarrow$ & $\delta^{p}\!\uparrow$ &
$\mathrm{Rel}^{p}\!\downarrow$ & $\delta^{p}\!\uparrow$ \\
\midrule
MoGe2 & 19.6 & 72.4 &
25.0 & 57.0 &
29.8 & 58.4 &
12.4 & 89.4 &
9.0 & 97.2 &
13.4 & 90.0 &
32.9 & 59.1 &
31.0 & 54.2 \\
\cdashline{1-17}
Mamba           & 8.4 & 93.5 & \cellcolor{custom_green!20}17.8 & \cellcolor{custom_green!20}77.1 & \cellcolor{custom_green!20}27.7 & 63.5 & 7.0 & 97.1 & \cellcolor{custom_green!70}5.9 & 98.1 & 10.1 & 91.0 & 24.0 & 60.0 & 25.1 & 66.5 \\
Trans. RNN      & \cellcolor{custom_green!20}6.7 & \cellcolor{custom_green!70}94.4 & 22.2 & 68.8 & 27.9 & \cellcolor{custom_green!20}64.5 & \cellcolor{custom_green!20}4.9 & \cellcolor{custom_green!20}98.8 & 7.6 & \cellcolor{custom_green!20}98.2 & \cellcolor{custom_green!20}9.3 & \cellcolor{custom_green!20}93.9 & \cellcolor{custom_green!20}15.7 & \cellcolor{custom_green!20}80.0 & \cellcolor{custom_green!20}17.3 & \cellcolor{custom_green!20}81.6 \\
Global Trans. & \cellcolor{custom_green!70}4.9 & \cellcolor{custom_green!20}94.2 &
\cellcolor{custom_green!70}10.1 & \cellcolor{custom_green!70}91.0 &
\cellcolor{custom_green!70}21.5 & \cellcolor{custom_green!70}75.6 &
\cellcolor{custom_green!70}2.1 & \cellcolor{custom_green!70}99.5 &
\cellcolor{custom_green!70}5.9 & \cellcolor{custom_green!70}99.0 &
\cellcolor{custom_green!70}8.8 & \cellcolor{custom_green!70}96.0 &
\cellcolor{custom_green!70}11.9 & \cellcolor{custom_green!70}89.1 &
\cellcolor{custom_green!70}9.7 & \cellcolor{custom_green!70}94.4 \\
\bottomrule
\end{tabular}
\end{threeparttable}
} 
\label{tab:supp_quanti_lr_stream_arch}
\end{subtable}

\vspace{2mm}

\begin{subtable}{0.8\textwidth}
\centering
\caption{\textbf{Effect of RoPE in the self-attention}.}
\small
\setlength{\tabcolsep}{4pt}
\begin{tabular}{lcccc}
\toprule
\multirow{2}{*}{Positional Embedding} &
\multicolumn{2}{c}{\textbf{Monkaa}~\cite{monkaa}} &
\multicolumn{2}{c}{\textbf{UrbanSyn}~\cite{urbansyn}} \\
& Rel$^p\!\downarrow$ & $\delta^{p}\!\uparrow$ & Rel$^p\!\downarrow$ & $\delta^{p}\!\uparrow$ \\
\midrule
None          & 11.0 & 89.9 & 10.1 & 93.9 \\
RoPE          & \textbf{9.7} &\textbf{ 92.1} & 10.3 & 93.5 \\
Interp. RoPE (ours)  & 10.1 & 91.0 & \textbf{8.8} & \textbf{96.0} \\
\bottomrule
\end{tabular}
\label{tab:supp_abl_rope_selfattn}
\end{subtable}

\vspace{2mm}

\begin{subtable}{0.8\textwidth}
\centering
\caption{\textbf{Effect of RoPE in the cross-attention}.}
\small
\setlength{\tabcolsep}{4pt}
\begin{tabular}{lcccc}
\toprule
\multirow{2}{*}{Positional Embedding} &
\multicolumn{2}{c}{\textbf{Monkaa}~\cite{monkaa}} &
\multicolumn{2}{c}{\textbf{UrbanSyn}~\cite{urbansyn}} \\
& Rel$^p\!\downarrow$ & $\delta^{p}\!\uparrow$ & Rel$^p\!\downarrow$ & $\delta^{p}\!\uparrow$ \\
\midrule
None          & 10.7 & \textbf{91.1} & 9.6 & 95.1 \\
``Snap'' RoPE (ours)         & \textbf{10.1} & 91.0 & \textbf{8.8} & \textbf{96.0} \\
\bottomrule
\end{tabular}
\label{tab:supp_abl_rope_crossattn}
\end{subtable}

\end{table*}

\subsection{More qualitative results}
\noindent\textbf{Interactive viewer (highly recommended).} The supplementary contains an HTML page (\texttt{webpage/index.html}) with side-by-side videos of predicted depth and reconstructed 3D pointmaps.

\begin{figure*}[t]
\begin{center}
\includegraphics[width=1.0\linewidth]{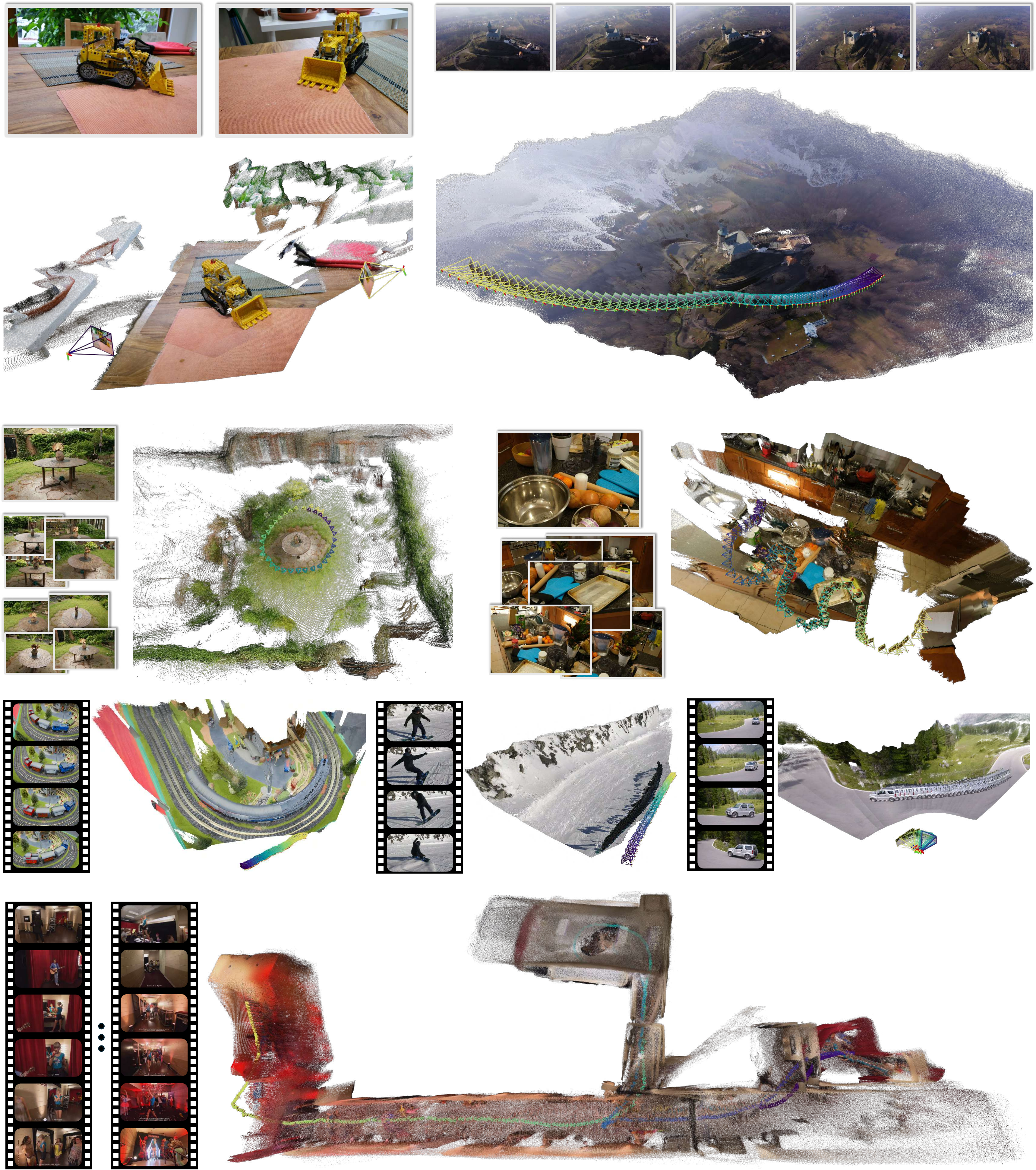}
\vspace{-2mm}
\caption{\textbf{Visualization of 3D pointmap reconstruction} on \textit{in-the-wild} scenarios.}
\vspace{-8mm}
\label{fig:supp_quali_pm}
\end{center}
\end{figure*}

Fig.~\ref{fig:supp_quali_pm} shows qualitative 3D pointmap reconstructions on in-the-wild scenes spanning static/dynamic motion, indoor/outdoor settings, and object-centric versus scene-level compositions.

\begin{figure*}[t]
\begin{center}
\includegraphics[width=1.0\linewidth]{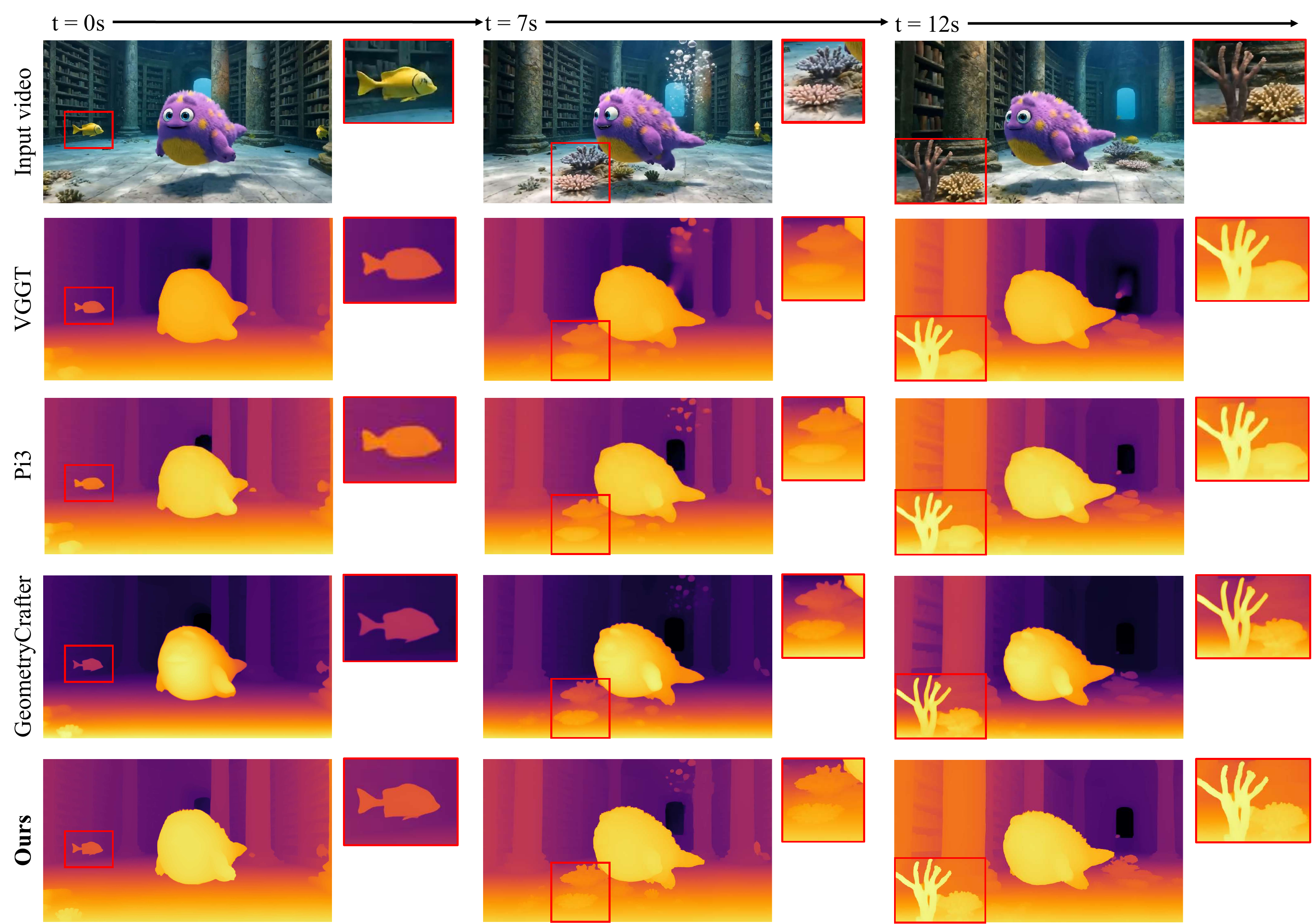}
\vspace{-2mm}
\caption{\textbf{Visualization of video depth estimation}. We compare our video depth prediction with VGGT~\cite{vggt}, Pi3~\cite{pi3}, and GeoemtryCrafter~\cite{geometrycrafter}. \Approach demonstrates more sharp and fine-grained predictions.}
\vspace{-8mm}
\label{fig:supp_quali_videodisp1}
\end{center}
\end{figure*}

\begin{figure*}[t]
\begin{center}
\includegraphics[width=1.0\linewidth]{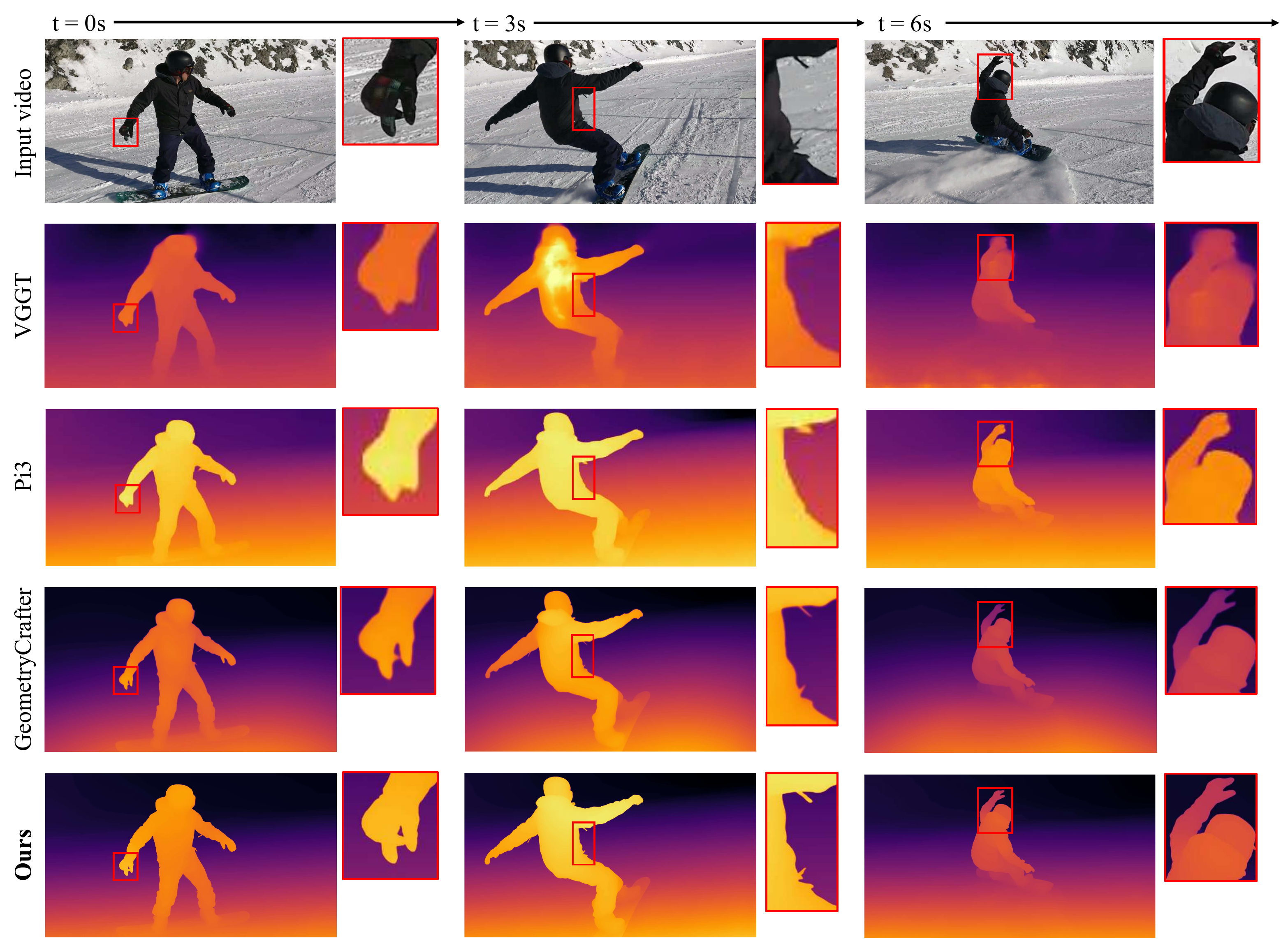}
\vspace{-2mm}
\caption{\textbf{Visualization of video depth estimation}. We compare our video depth prediction with VGGT~\cite{vggt}, Pi3~\cite{pi3}, and GeoemtryCrafter~\cite{geometrycrafter}. \Approach demonstrates more sharp and fine-grained predictions.}\vspace{-8mm}
\label{fig:supp_quali_videodisp2}
\end{center}
\end{figure*}

\begin{figure*}[t]
\begin{center}
\includegraphics[width=1.0\linewidth]{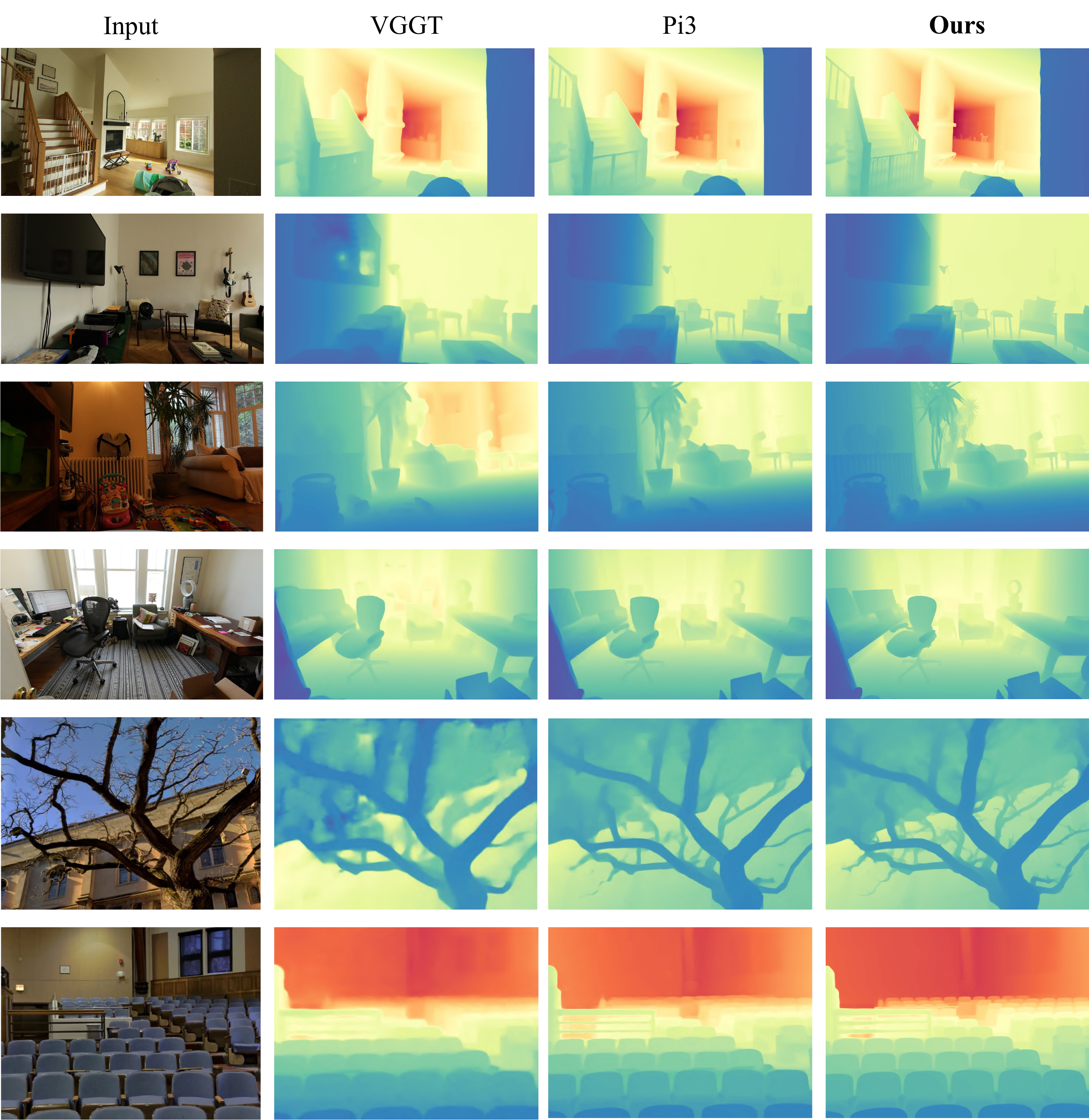}
\vspace{-2mm}
\caption{\textbf{Visualization of depth estimation} on static scenes.}
\vspace{-8mm}
\label{fig:supp_quali_depthmap}
\end{center}
\end{figure*}

Figs.~\ref{fig:supp_quali_videodisp1},~\ref{fig:supp_quali_videodisp2},~\ref{fig:supp_quali_depthmap} compare our video depth to recent state-of-the-art methods~\cite{vggt,pi3,geometrycrafter}, highlighting sharper boundaries and stronger temporal stability.

\begin{figure*}[t]
\begin{center}
\includegraphics[width=1.0\linewidth]{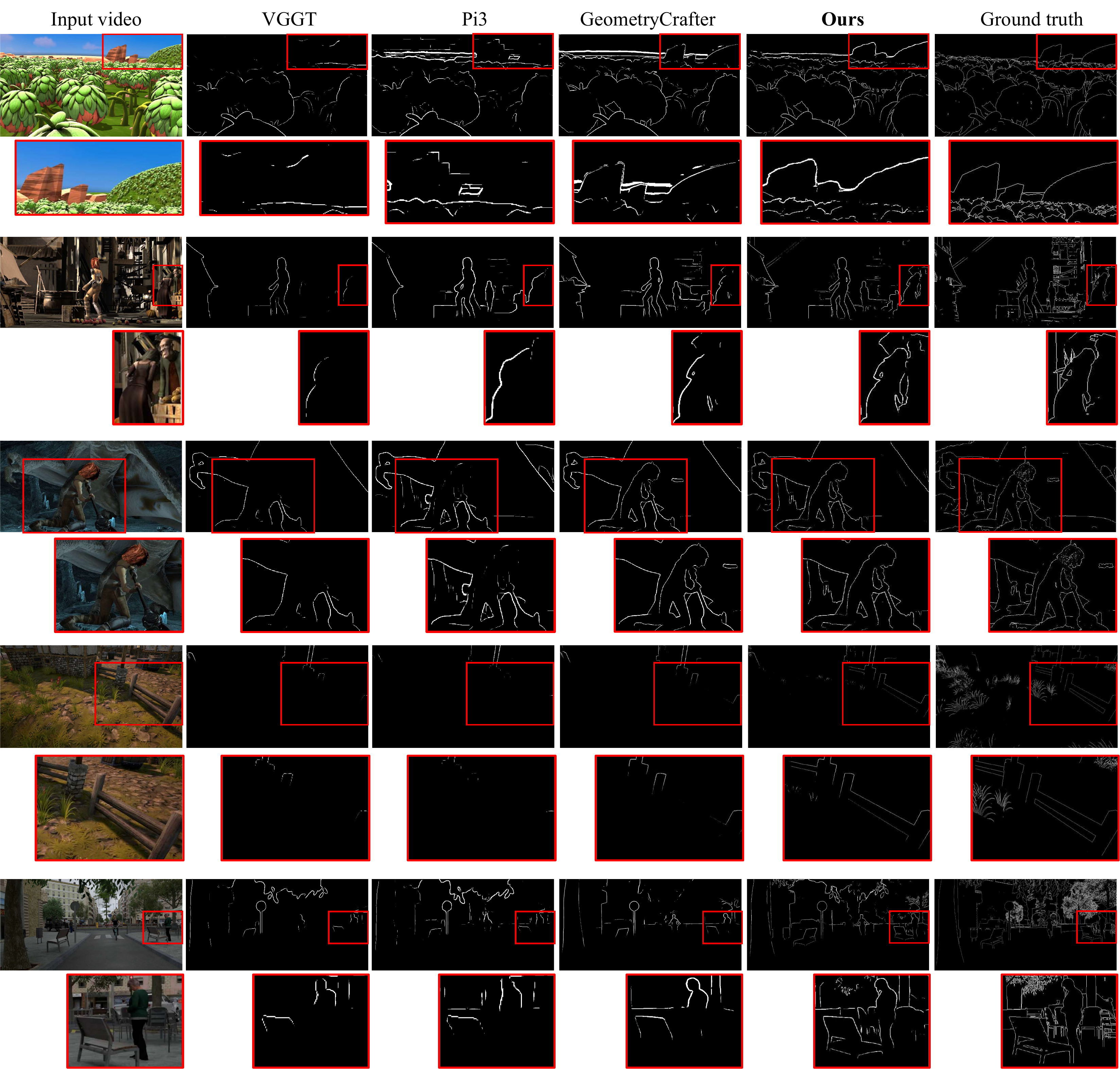}
\vspace{-2mm}
\caption{\textbf{Visualization of predicted depth edge maps}, which are defined by a depth ratio between neighboring pixels above a threshold. We zoom-in the edge map details in the \textcolor{red}{red bounding boxes}.}
\vspace{-8mm}
\label{fig:supp_quali_edgemap}
\end{center}
\end{figure*}

Fig.~\ref{fig:supp_quali_edgemap} visualizes depth-edge maps—the contours obtained by thresholding neighboring-pixel depth changes. Compared to baselines~\cite{vggt,pi3,geometrycrafter}, our results capture thin structures and small or distant objects more reliably.

\begin{figure*}[t]
\begin{center}
\includegraphics[width=1.0\linewidth]{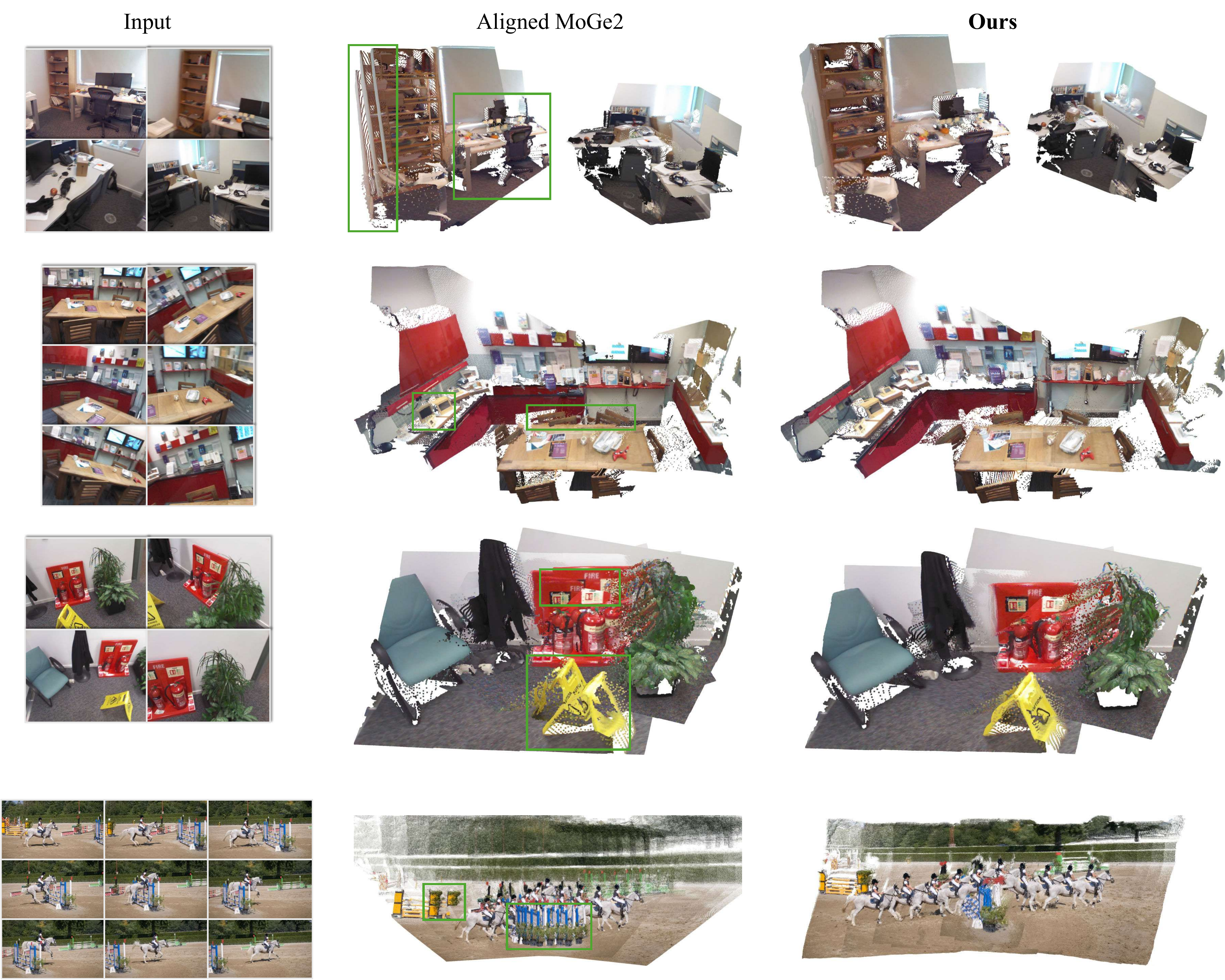}
\vspace{-2mm}
\caption{\textbf{Predicted 3D pointmaps} of the aligned MoGe2 baseline and our method. The aligned MoGe2 baseline exhibits layering artifacts (\textcolor[RGB]{78,167,46}{green boxes}) due to the lack of strong multi-view binding.}
\vspace{-8mm}
\label{fig:supp_compare_alignmoge2}
\end{center}
\end{figure*}

Fig.~\ref{fig:supp_compare_alignmoge2} compares 3D pointmaps from \Approach to an \emph{aligned-MoGe2} baseline. In Tab. 6 (Sec. 4.6), we define \textbf{Setting A}: run MoGe2~\cite{moge2} per frame and \emph{post hoc} align each predicted pointmap to a globally consistent pointmap from Pi3~\cite{pi3}. This simple alignment recovers fine detail and enforces a shared scale, but—as the figure shows—still produces layering/stitching artifacts because depth is estimated independently per frame without strong cross-view coupling.

\begin{figure*}[t]
\begin{center}
\includegraphics[width=1.0\linewidth]{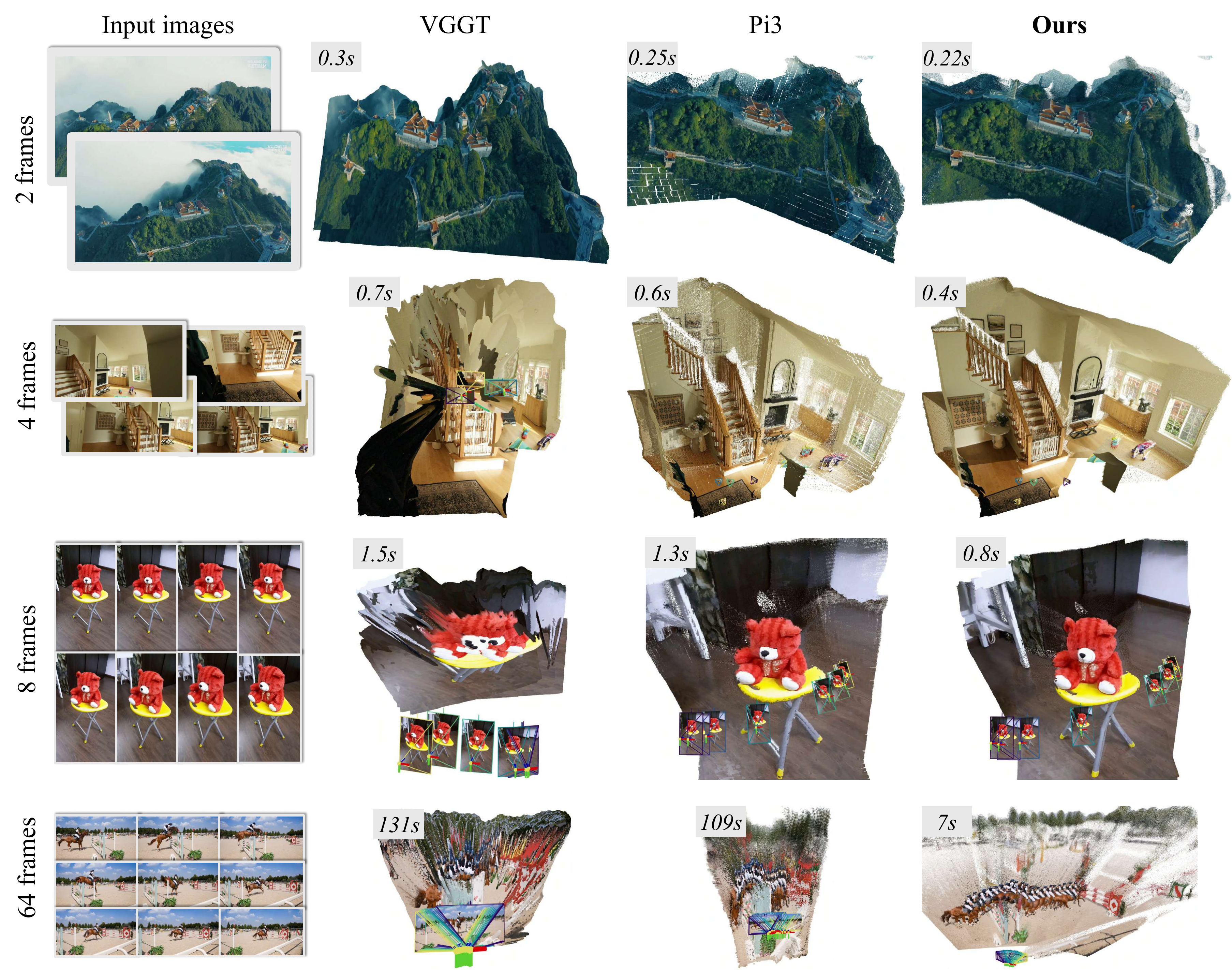}
\vspace{-2mm}
\caption{\textbf{Visualization of 3D reconstruction} with high-resolution inputs.}
\vspace{-8mm}
\label{fig:supp_quali_hr_pointmap}
\end{center}
\end{figure*}

Fig.~\ref{fig:supp_quali_hr_pointmap} visualizes 3D pointmaps reconstructed from 2K inputs. \Approach runs substantially faster—especially on longer clips—while producing more plausible, multi-view–consistent reconstructions. In contrast, global-attention baselines~\cite{pi3,vggt} either run out of memory or degrade at this resolution.

\section{High-resolution inference analysis of visual-geometry models}
We analyze how pretrained feed-forward visual-geometry models~\cite{vggt,pi3} behave when evaluated well beyond their training resolution (up to 2K on the long side).

\begin{figure*}[t]
\centering

\begin{subfigure}{\linewidth}
  \centering
  \includegraphics[width=0.77\linewidth]{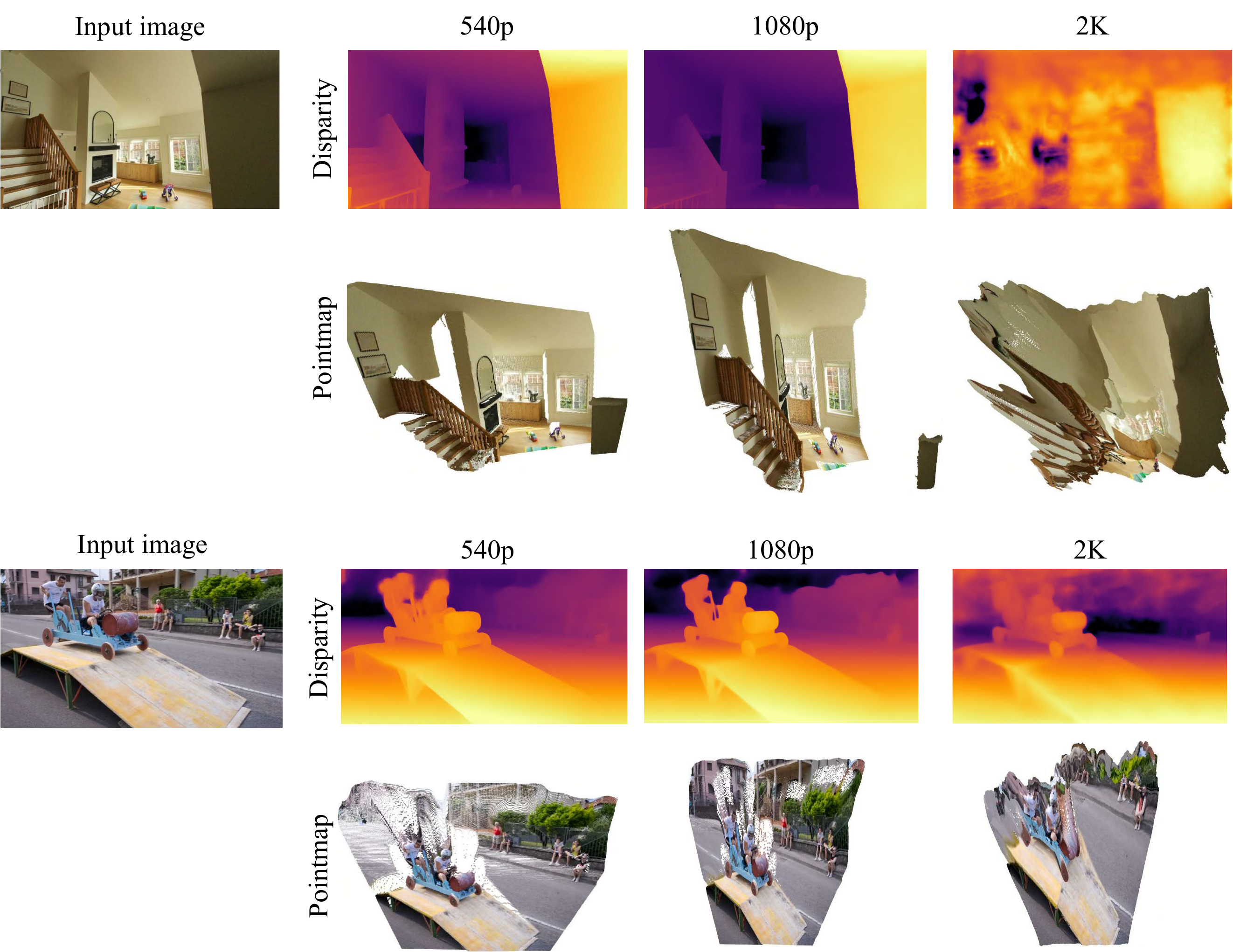}
  \caption{\textbf{High-resolution single-image inference} of VGGT~\cite{vggt}}
  \label{fig:quali_singleimg_vggt}
\end{subfigure}

\vspace{2mm}

\begin{subfigure}{\linewidth}
  \centering
  \includegraphics[width=0.77\linewidth]{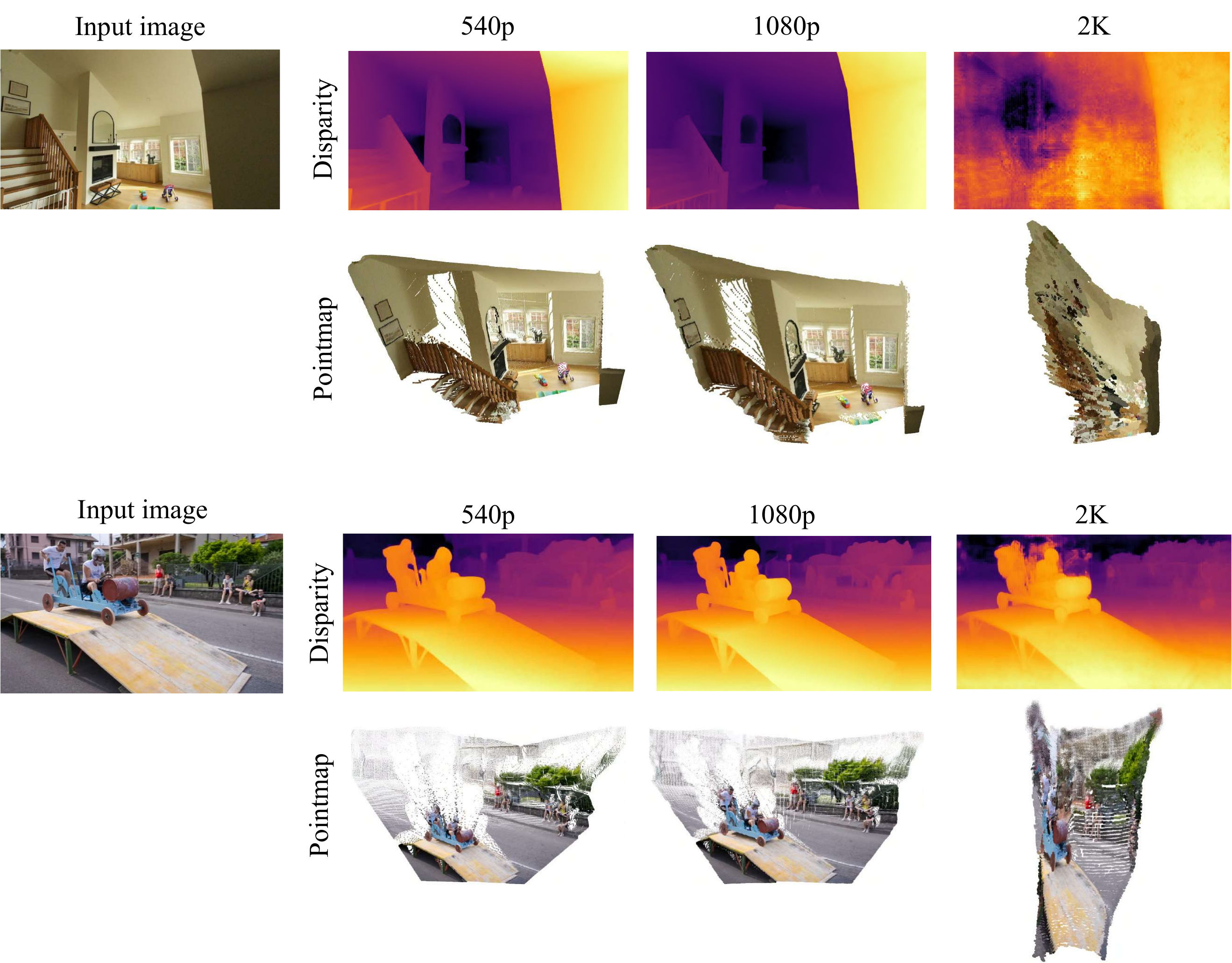}
  \caption{\textbf{High-resolution single-image inference} of Pi3~\cite{pi3}}
  \label{fig:quali_singleimg_pi3}
\end{subfigure}

\vspace{-2mm}
\caption{Qualitative results for high-resolution single-image inference: (a) VGGT~\cite{vggt} and (b) Pi3~\cite{pi3}.}
\label{fig:quali_singleimg_both}
\vspace{-8mm}
\end{figure*}

\myheading{Single-image stress test.}
We resize single-image inputs to several resolutions (e.g., 540p, 1080p, and 2K) and run the public checkpoints of VGGT~\cite{vggt} and Pi3~\cite{pi3} without any architectural changes. We visualize depth maps and corresponding 3D pointmaps (VGGT in Fig.~\ref{fig:quali_singleimg_vggt}, Pi3 in Fig.~\ref{fig:quali_singleimg_pi3}).

\noindent
At \(\sim\!540\)p, both methods produce plausible geometry. When the resolution is increased to \(\sim\!1080\)p, predictions exhibit shape distortions; at \(\sim\!2\)K, outputs often collapse into fragmented or globally inconsistent pointmaps. These failures are consistent across scenes.

\begin{figure*}[t]
\begin{center}
\includegraphics[width=0.95\linewidth]{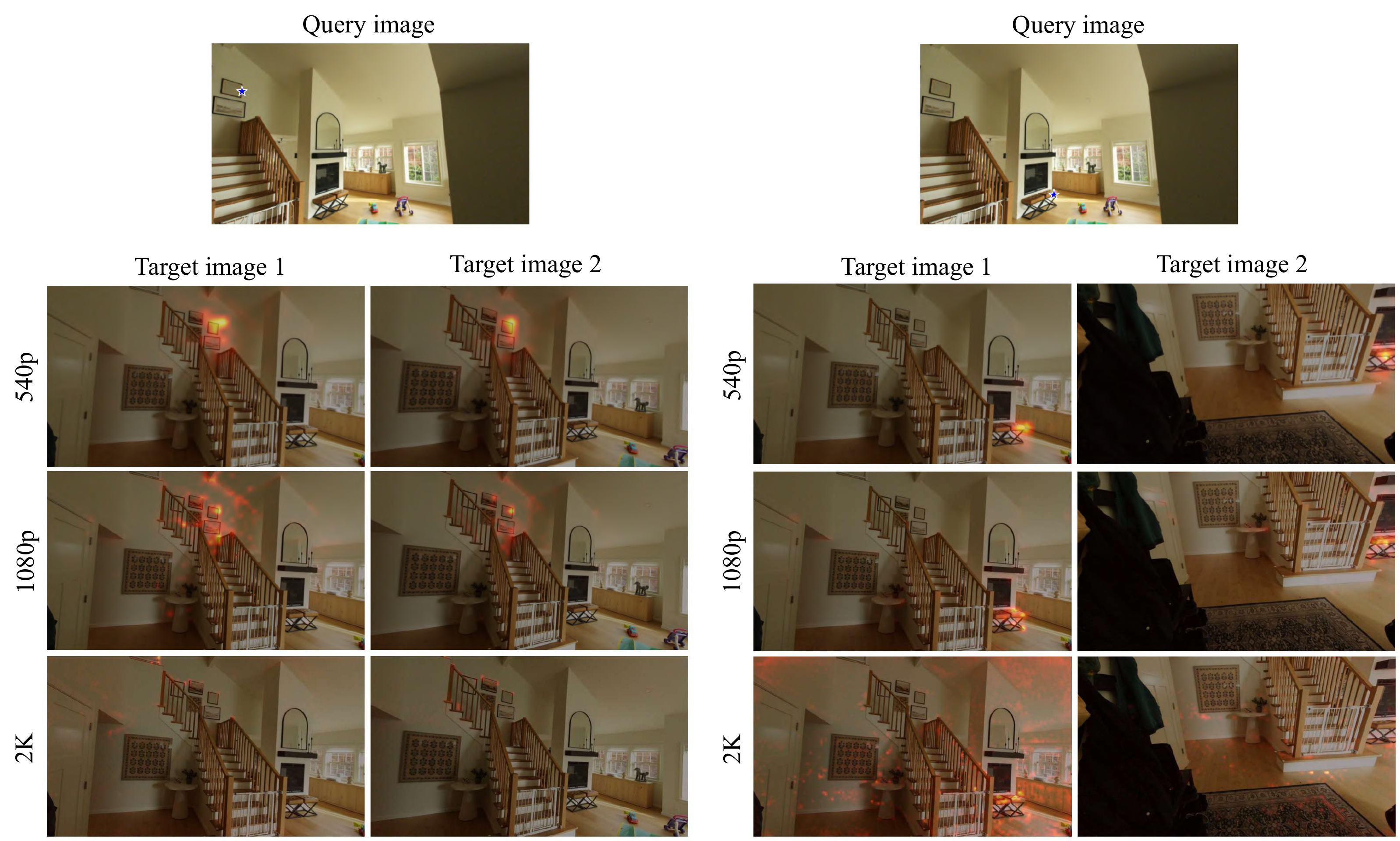}

\vspace{-2mm}
\caption{\textbf{Attention map} of the 15th global-attention layer of VGGT~\cite{vggt} at different input resolutions. The query token in the first image is marked with a \textcolor{blue}{blue star}.}

\vspace{-8mm}
\label{fig:vggt_attn1}
\end{center}
\end{figure*}
\begin{figure*}[t]
\begin{center}
\includegraphics[width=0.95\linewidth]{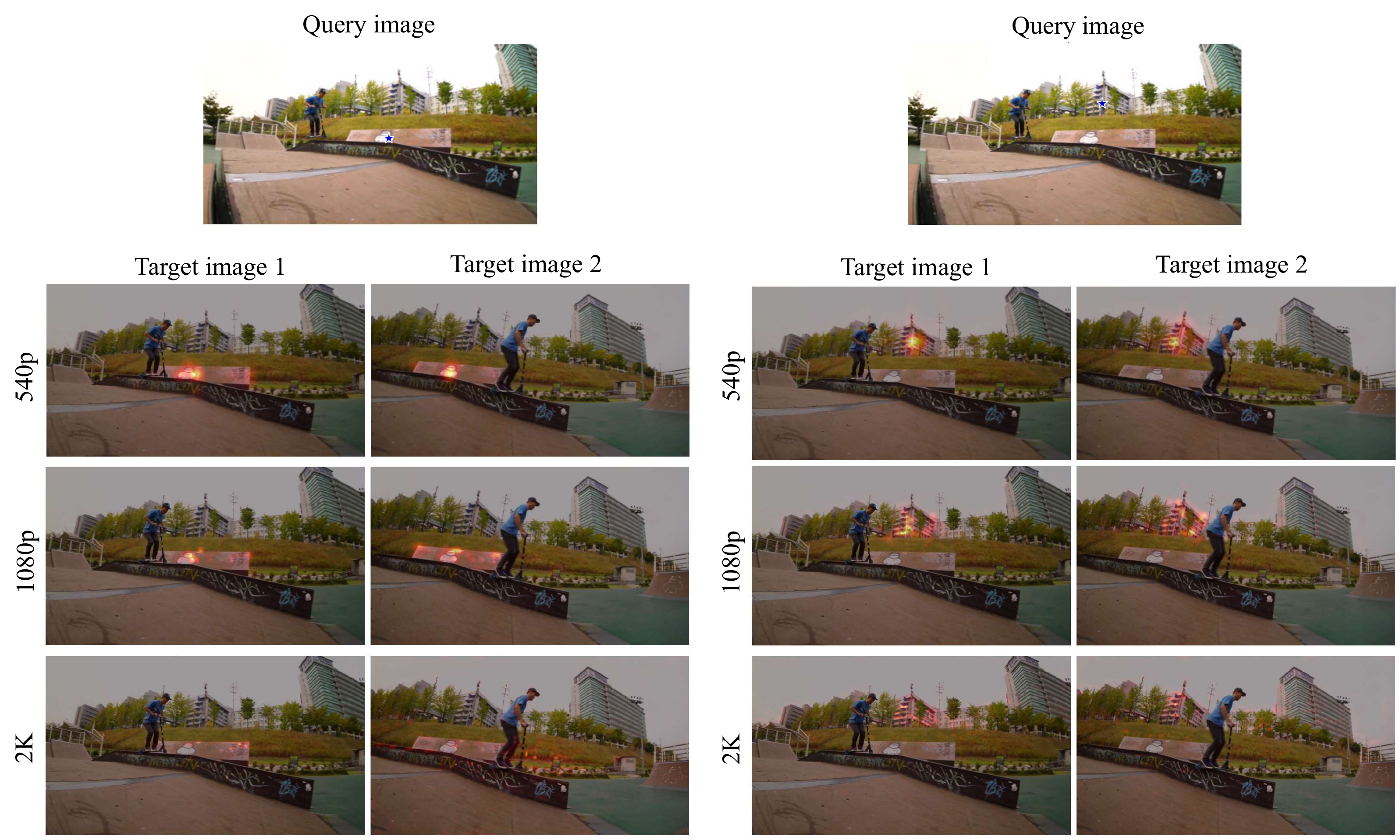}

\vspace{-2mm}
\caption{\textbf{Attention map} of the 15th global-attention layer of VGGT~\cite{vggt} at different input resolutions. The query token in the first image is marked with a \textcolor{blue}{blue star}.}
\vspace{-8mm}
\label{fig:vggt_attn2}
\end{center}
\end{figure*}
\begin{figure*}[t]
\begin{center}
\includegraphics[width=0.95\linewidth]{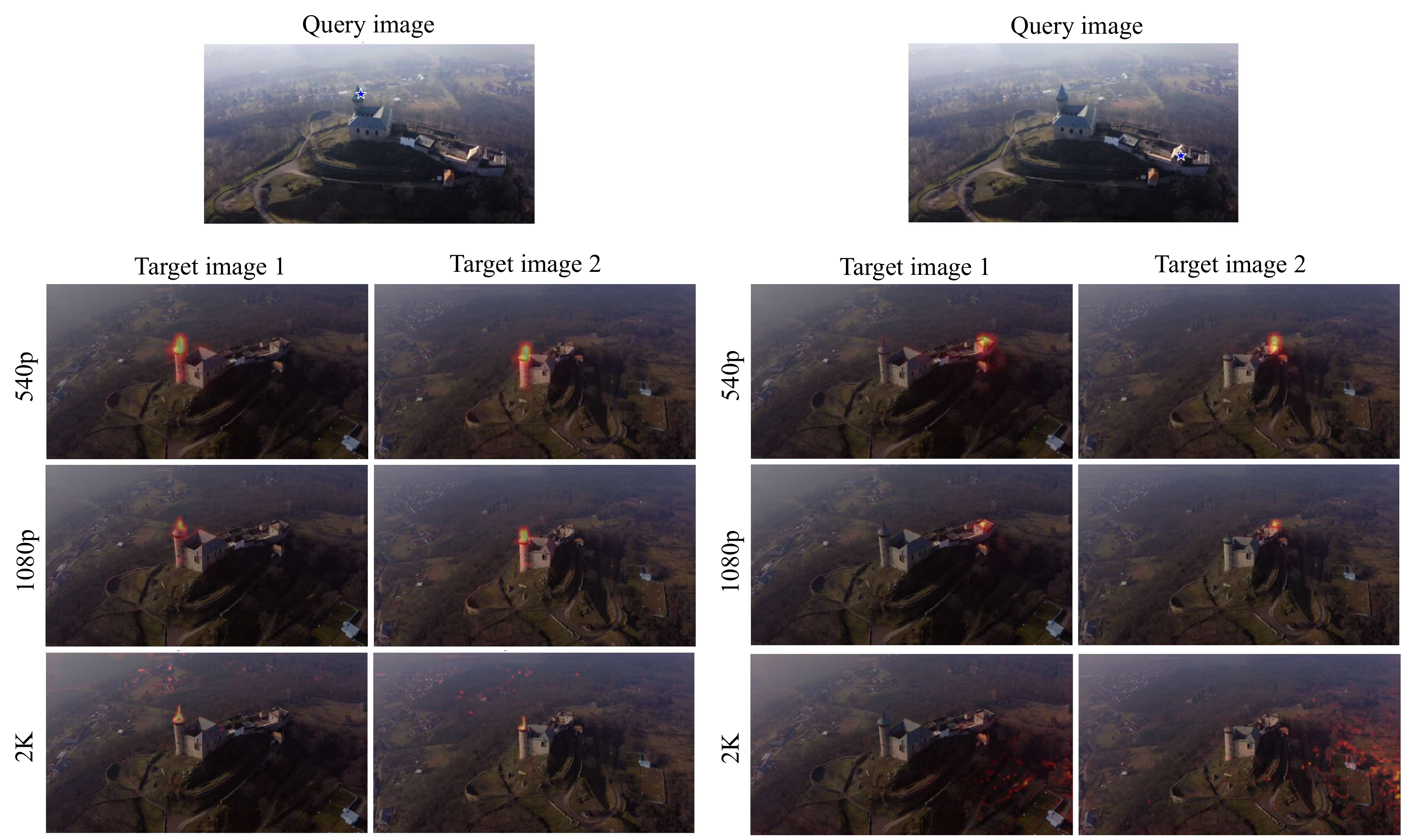}
\vspace{-2mm}
\caption{\textbf{Attention map} of the 15th global-attention layer of VGGT~\cite{vggt} at different input resolutions. The query token in the first image is marked with a \textcolor{blue}{blue star}.}
\vspace{-8mm}
\label{fig:vggt_attn3}
\end{center}
\end{figure*}

\myheading{Global-attention behavior (3-view input).}
To probe failure modes under high resolution, we evaluate VGGT with \emph{triplets} of views (3 frames), not single images. We fix the number of views to three and vary only the spatial resolution. Following prior observations that global layers perform exhaustive correspondence search~\cite{fastervggt}, we visualize post-softmax maps for a few query tokens in view~1 and overlay their responses in the other views (Figs.~\ref{fig:vggt_attn1}–\ref{fig:vggt_attn3}). At \(\sim\!540\)p, the maps are compact and centered on true correspondences. As resolution increases, attention becomes diffuse and multi-modal, drifting toward semantically similar yet geometrically incorrect regions; by 2K it degenerates into high-entropy responses with no clear matches.

\myheading{Likely causes.}
(i) \textit{Positional extrapolation:} standard rotary/absolute positional parameterizations learned at \(\sim\)540 px do not extrapolate reliably to much larger token grids, skewing query–key phases and degrading similarity scores~\cite{rope_interp}.
(ii) \textit{Entropy growth:} increasing resolution raises token count without increasing the effective receptive field, making correspondence sparser per token and increasing attention entropy~\cite{jin2023training}.  
(iii) \textit{Distribution shift:} training rarely exposes models to high-frequency, high-resolution statistics; the learned global matcher thus overfits to lower-res aliasing patterns.

From our experiments, we find that naively scaling input resolution is unreliable for current global-attention pipelines: at 1K–2K, pretrained models often exhibit correspondence collapse—diffuse attention and distorted depth/pointmaps—likely due to positional-encoding extrapolation and distribution shifts. Therefore, in our proposed ~\Approach, we amortize global aggregation at low resolution and fuse it into a per-frame high-resolution path; this preserves detail at 2K while keeping memory and runtime practical. Furthermore, to stabilize high-res inference, we adopt resolution-aware positional encodings (interpolated RoPE), explicit cross-scale alignment (snapping HR token coordinates to the LR grid for cross-attention), and multi-scale training that includes high-res regimes.

\end{document}